\documentclass{article}

\usepackage[main, final]{neurips_2026}

\usepackage[utf8]{inputenc}
\usepackage[T1]{fontenc}
\usepackage{hyperref}
\usepackage{url}
\usepackage{booktabs}
\usepackage{amsmath}
\usepackage{amssymb}
\usepackage{amsfonts}
\usepackage{nicefrac}
\usepackage{microtype}
\usepackage{xcolor}
\usepackage{graphicx}
\usepackage{wrapfig}
\usepackage{caption}
\usepackage{algorithm}
\usepackage{algorithmic}
\usepackage{subcaption}

\title{LIPAR: Training-free Latent Inter-Frame Pruning with Attention Recovery}

\hypersetup{
  colorlinks=true,
  linkcolor=black,
  citecolor=black,
  urlcolor=blue
}

\author{%
  Dennis Menn$^{1}$\thanks{Correspondence to: \texttt{dennismenn@utexas.edu}} \quad Yuedong Yang$^{1}$ \quad Bokun Wang$^{1}$ \quad Xiwen Wei$^{1}$ \quad Mustafa Munir$^{2}$ \\[0.5ex]
  \textbf{Feng Liang$^{3}$ \quad Radu Marculescu$^{1}$ \quad Chenfeng Xu$^{1}$ \quad Diana Marculescu$^{1}$} \\[0.8ex]
  $^{1}$University of Texas at Austin \quad $^{2}$NVIDIA \quad $^{3}$Meta \\[0.8ex]
  \textbf{Project Page:} \href{https://dennismenn.github.io/lipar}{\texttt{dennismenn.github.io/lipar}} \quad \textbf{Code:} \href{https://github.com/DennisMenn/lipar}{\texttt{github.com/DennisMenn/lipar}}
}

\begin{document}

\maketitle

\begin{center}
    \vspace{-2em}    \centerline{\includegraphics[width=1\linewidth]{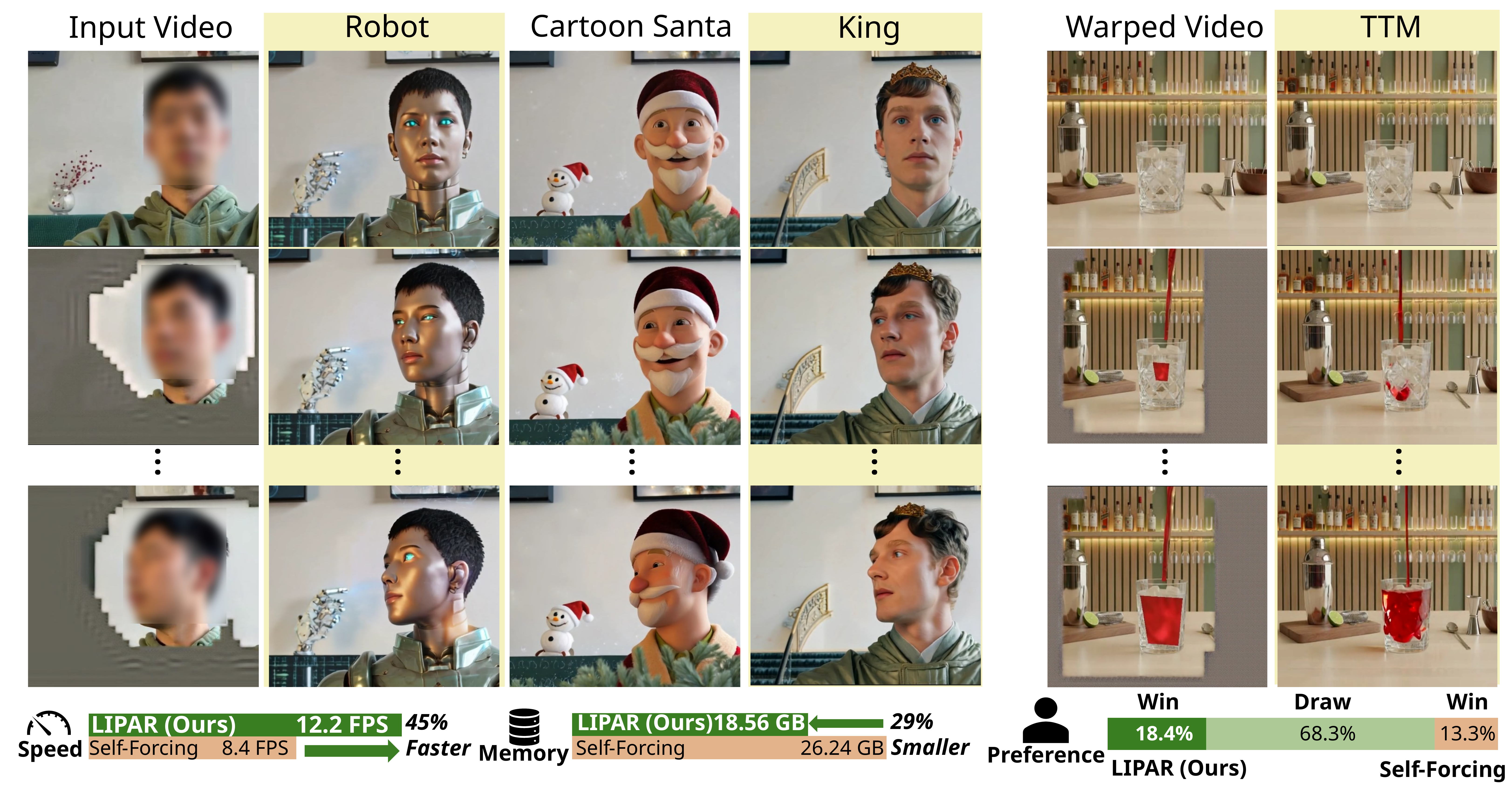}}
    \vspace{-1em}
    \captionof{figure}{We present LIPAR, a pruning framework that increases inference speed and reduces GPU memory usage for conditioned video generation by reusing temporally unchanged latents (gray regions). With our proposed Attention Recovery, LIPAR integrates with the Diffusion Transformer seamlessly in a training-free manner without degrading visual quality.}   
  
    \label{fig:placeholder}
  \end{center}

\vspace{\dimexpr-1em-1.3ex\relax}

\begin{abstract}
Video generation enables text-to-video synthesis, video editing, and motion-controlled content creation. However, current video generation models suffer from high computational latency, rendering true real-time capabilities infeasible for downstream tasks. We address this limitation by exploiting the temporal redundancy inherent in video latent patches. To this end, we propose the \textbf{Latent Inter-frame Pruning with Attention Recovery (LIPAR)} framework, which detects and skips recomputing duplicated latent patches. Additionally, we introduce a novel \textbf{Attention Recovery} mechanism that approximates the attention values of pruned tokens, thereby removing visual artifacts arising from naively applying the pruning method. Empirically, our method increases generation throughput by $1.45\times$, on average achieving 12.2 FPS on an NVIDIA A6000 compared to the baseline 8.4 FPS. The proposed method does not compromise generation quality and can be seamlessly integrated with Diffusion Transformer without additional training. Our approach effectively bridges the gap between traditional compression algorithms and modern generative pipelines.
\end{abstract}

\section{Introduction}
\label{sec:intro}
Image and video synthesis have emerged as transformative technologies in generative AI, reshaping fields from film production to robotics \citep{realwonder2026, gao2026dreamdojo}. In particular, Diffusion Transformers (DiTs) have emerged as a dominant force in generative tasks, achieving remarkable success in high-fidelity image and video synthesis~\citep{Peebles2022DiT, kong2024hunyuanvideo}. However, their practical deployment is severely constrained by computational inefficiencies~\citep{yang2025sparse, xi2025sparse}. Despite recent advances, such as causal attention adaption and few-step distillation~\citep{yin2025causvid}, video generation remains a compute-demanding task. Furthermore, high computation costs impede real-time human-machine interaction (\textit{e.g.}, defined by achieving 30 fps) on a single GPU~\citep{feng2025streamdiffusionv2, shin2025motionstream, singer2025timetomove, huang2025selfforcing}.

To reduce computational costs, traditional video compression algorithms identify repeated patches in temporal and spatial dimensions to avoid reprocessing them in pixel space ~\citep{MPEG1991}. In contrast, the current Latent Diffusion Model (LDM) framework allocates fixed compute for every token, regardless of redundancy in the content~\citep{rombach2021ldm, kong2024hunyuanvideo, wan2025}. This is primarily due to the limited understanding of semantics in the latent space and the difficulty in pinpointing redundancy prior to the generation process.

Previous methods attempted to exploit redundancy by merging similar tokens within each attention block to prevent re-computation~\citep{bolya2022tome, bolya2023tomesd, wu2025importancetome, fang2025attend}. However, these methods suffer from computational overhead due to the frequent, expensive process of determining similar tokens for each block (bipartite matching); additionally, token merging is restricted to certain layers, thereby failing to save computation across all layers~\citep{bolya2023tomesd, wu2025importancetome, fang2025attend} and may result in visual artifacts in the causal attention backbone~\citep{huang2025selfforcing} as shown experimentally.

To bridge the gap between traditional video compression and modern LDM frameworks, we propose \textbf{Latent Inter-frame Pruning with Attention Recovery (LIPAR)} for conditioned video generation. We first observe that a strong correlation between the pixel-space and latent-space dynamics allowing to bridge the gap between traditional video compression and current LDM techniques. This finding allows us to bypass expensive bipartite matching and instead use an inexpensive $L_1$ difference to detect redundant patches in the latent space. Furthermore, we find that naively pruning tokens introduces visual artifacts in the generated videos due to the resulting train-inference misalignment. To close this gap, we propose \textbf{Attention Recovery}, which lifts the restriction in prior methods that merged tokens must be unmerged after a fixed number of layers, enabling the speedup to extend across all layers while substantially improving output quality.

When tested on 222 video-text prompts from the Davis~\citep{davis2017dataset} and OpenHumanVid~\citep{li2024openhumanvid} datasets, LIPAR achieves more than $1.45\times$ speedup with a $29\%$ reduction in GPU usage (requiring only $18.6$ GB). Rigorous human evaluation yields an $86.7\%$ win-tie rate against the original (unpruned) results, demonstrating high visual quality and a clear improvement over existing training-free pruning methods. Additionally, our method generalizes from causal attention~\citep{yin2025causvid, huang2025selfforcing} to bidirectional attention~\citep{wan2025}. Our contributions are summarized as follows:
\begin{enumerate}
    \item We identify \textbf{strong Pearson correlations} between pixel-space and latent-space dynamics, motivating the adaptation of traditional video compression algorithms to the generative pipeline (Section~\ref{sec:empirical_evidence}).
    \item We provide a \textbf{theoretical formulation} of the training-inference discrepancy induced by direct token pruning and establish a general mathematical condition that pruning scheme must satisfy to preserve visual quality (Section~\ref{sec:proble_formulation}).
    \item We design a pipeline that integrates \textbf{inter-frame compression} with LDMs for video editing, bridging traditional video compression and modern generative models (Section~\ref{sec:LIPAR_overview}).
    \item To mitigate the train-inference misalignment caused by pruning, we propose \textbf{Attention Recovery}, which approximates the output of the unpruned token sequence. This enables training-free integration while preserving visual quality, achieves an $O(n)$ speedup (where $n$ is the number of retained tokens), and generalizes to both causal and bidirectional attention (Section~\ref{sec:attention_recovery}).
    
\end{enumerate}

\section{Related Work}
\label{apx:related_works}
\paragraph{\textbf{Accelerating Diffusion Models}}
To mitigate the high computational cost of Video Diffusion models, sparse video generation methods~\citep{xi2025sparse, yang2025sparse} optimize attention computation through semantic-aware permutation techniques. Another line of research accelerates generation via feature caching, skipping specific layers across denoising timesteps~\citep{kahatapitiya2025adaptive, liu2024timestep}. Additionally, CausVid~\citep{yin2025causvid} applies few-step distillation~\citep{yin2024improved} to accelerate video generative models. Orthogonal to these techniques, we save computation by reducing the number of input tokens, exploiting the temporal redundancy inherent in the latent space.

Several methods reduce computation by merging tokens within selected layers at inference time. \citet{bolya2022tome, bolya2023tomesd} introduced a bipartite matching algorithm to merge redundant tokens in Transformers, and subsequent work refined the selection criterion to prioritize semantically important tokens~\citep{fang2025attend, li2023vidtome, wu2025importancetome}. In contrast, LIPAR enables end-to-end pruning across all layers via Attention Recovery and replaces expensive token-selection procedures with fast inter-frame pruning.


\paragraph{\textbf{Real-time Conditioned Video Generation}}
\label{appendix:related_works}
Recent advances in video generation have shifted toward interactivity, paving the way for \textit{real-time, conditionally generated video} that responds instantly to user input. We focus on two prominent tasks: Real-time Video Editing \citep{Munir_2026_WACV, flynn2026edityourself, kodaira2025streamdit, feng2025streamdiffusionv2, kodaira2023streamdiffusion} and Motion Control \citep{luo2026shotstream, realwonder2026, chhatre2026trajectory, shin2025motionstream, singer2025timetomove}. Real-time video editing targets live applications, providing instantaneous edits based on user prompts. While approaches leveraging few-step image diffusion models achieve low latency on consumer-grade GPUs~\citep{kodaira2023streamdiffusion, liang2024looking, Munir_2026_WACV}, maintaining temporal consistency remains challenging. In contrast, \citep{feng2025streamdiffusionv2} adapts video diffusion models for real-time editing, yet computational costs still hinder single-GPU performance.

Motion Control guides synthesis via explicit motion signals. Recently, \citet{shin2025motionstream} and \citet{singer2025timetomove} introduced techniques to generate motion-conditioned videos using warped static image enabling intuitive interactions, such as dragging a dog's head to turn~\citep{shin2025motionstream, singer2025timetomove, realwonder2026, chhatre2026trajectory}. However, achieving real-time (30 FPS) response on a consumer-grade GPU remains challenging.

\section{Motivation: Empirical Evidence}
\label{sec:empirical_evidence}
A fundamental principle of video compression in pixel space is that redundant pixels need not be re-transmitted~\citep{MPEG1991}. To adapt this idea from pixel to latent space, the latent space must also contain redundancy; if so, we can copy redundant patches instead of re-generating them, saving computation. A straightforward way to test this is to check whether pixel-level temporal dynamics are preserved within the latent manifold. \textbf{We measure the correlation between changes in pixel space and changes in latent space along the temporal axis.} A strong correlation indicates that a patch unchanged in pixel space will remain unchanged in latent space:
\begin{equation}
    \mathrm{Corr}\Bigl( \lVert p^{(t,x,y)}_\mathrm{pixel} - p^{(t+1,x,y)}_\mathrm{pixel}\rVert_1, \quad \lVert p^{(t,x,y)}_\mathrm{latent} - p^{(t+1,x,y)}_\mathrm{latent}\rVert_1 \Bigr)
    \label{eqn:pearson}
\end{equation}
where $p$ is a patch in the pixel or latent space, and $(t,x,y)$ denotes its spatial and temporal position. We conducted this analysis on the entire DAVIS 2017 train-val set \citep{davis2017dataset}, using a latent patch size of $(2, 2, 2)$ across the temporal and spatial axes. To ensure generalizability, we tested both the WAN 2.1 VAE and WAN 2.2 TI2V VAE \citep{wan2025}. We employ the $L_1$ norm to quantify change and Pearson correlation, as in Eqn. \ref{eqn:pearson}, to measure the relationship between the two spaces.

Our results show a strong correlation between pixel-space and latent-space changes: 0.69 for WAN 2.1 VAE and 0.77 for WAN 2.2 VAE. Note that given the heavy spatial compression performed by the attention layers in the encoder, there is no \textit{a priori} guarantee that the latent manifold would be preserved in the raw pixel space.

\begin{wrapfigure}[10]{r}{0.5\linewidth}
    \centering
    \vspace{-5em}
    \includegraphics[width=1\linewidth]{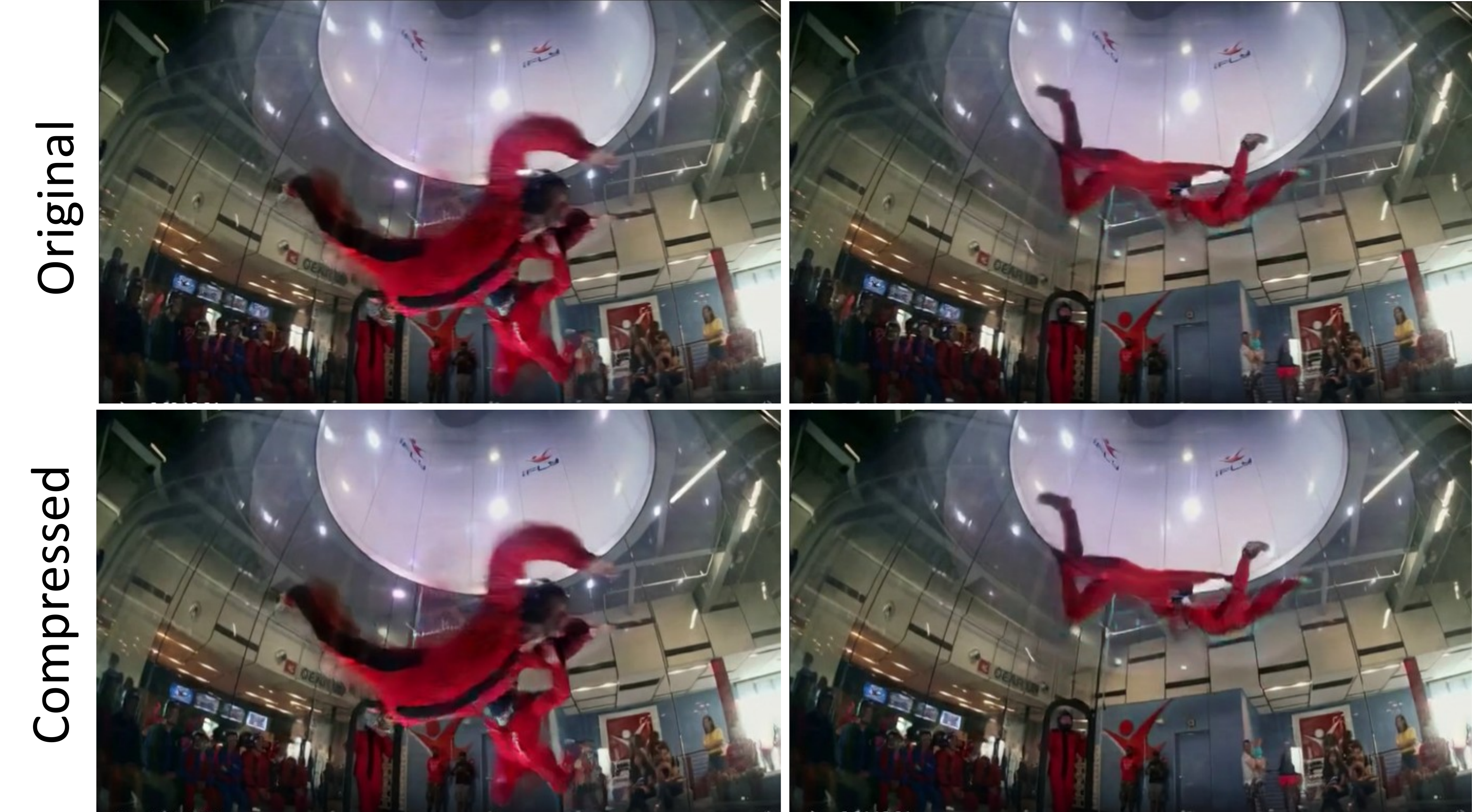}
    \caption{\textbf{Decoding Compressed Latents.} \textit{Original}: direct decoding. \textit{Compressed}: replaced patches with previous-frame counterparts.}
    \label{fig:compress_latents}

\end{wrapfigure}

To test temporal redundancy in the latent space, we select ten videos from the DAVIS dataset and substitute (nearly) unchanged patches with those from the previous frame to create a ``compressed'' latents. Even after compressing 46\% of the latents, the decoded output resembled the originally decoded video, with an averaged Learned Perceptual Image Patch Similarity score (LPIPS) $\leq 0.05$ \citep{zhang2018perceptual}, as shown in Figure \ref{fig:compress_latents}. For the detailed experimental settings, please refer to Appendix \ref{appedix:latents_compression}. These findings confirm that temporal redundancy exists in the latent space and support the adaptation of video compression methods to the latent space.

\section{Problem Formulation}
\label{sec:proble_formulation}
\subsection{Target Objective}
The goal of this approach is to use the generated tokens of the temporally pruned sequence to approximate those of the full, unpruned sequence. Formally, we require the reconstructed output, obtained by pruning, denoising, and then duplicating, to approximate the original denoised output, as shown on the left side of Eqn.~\ref{eqn:goal}, where $x_t$ is the token sequence at time $t$, $\mathcal{P}$ is the pruning operator, $D$ is the denoising network, and $\mathcal{R}$ is the recovery operator (reusing temporal predecessors).

If we restrict pruning to temporally redundant tokens, the left side of Eqn.~\ref{eqn:goal} simplifies to the right side. Since the pruned tokens are close in value to their predecessors, we only need to ensure that the kept tokens approximate those of the full sequence. Our goal thus simplifies to ensuring that the denoising operation \textit{commutes} with the pruning operation:
\begin{equation}
    \underbrace{\mathcal{R}\big(D(\mathcal{P}(x_t))\big) \approx D(x_t)}_{\text{Goal}} 
    \implies 
    D(\mathcal{P}(x_t)) \approx \mathcal{P}(D(x_t))
\label{eqn:goal}
\end{equation}

 

Note that the sufficient condition for Eqn.~\ref{eqn:goal} to hold reduces to approximating the Multi-head Self-Attention (MSA) outputs of the unpruned sequence from the pruned sequence within each block, as shown in Eqn.~\ref{eqn:msa_equivalence} below. This is because self-attention is the only operation that depends on the entire token sequence; if its outputs are approximated, subsequent per-token operations (\textit{e.g.}, cross-attention and linear layers) will align correspondingly, preserving the overall approximation. See Appendix~\ref{appxpendix: Derive Target Objective} for the derivation.
\begin{equation}
    \operatorname{MSA}(\mathcal{P}(x_t)) \approx \mathcal{P}(\operatorname{MSA}(x_t))
    \label{eqn:msa_equivalence}
\end{equation}

\subsection{MSA Approximation}
\begin{wrapfigure}{r}{0.55\linewidth}
    \centering
    \includegraphics[width=\linewidth]{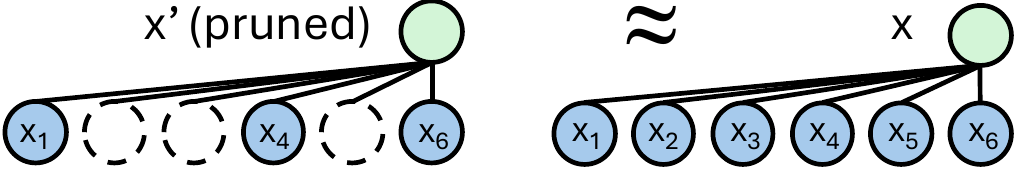}
    \caption{Approximation of pruned tokens to the unpruned token sequence. Dashed circles indicate pruned tokens, where $x_1 \approx x_2 \approx x_3$ and $x_4 \approx x_5$.}
    \vspace{-1em}
    \label{fig:math_goal}
\end{wrapfigure}

To satisfy Eqn. \ref{eqn:msa_equivalence}, we consider the one-dimensional case where tokens have fixed spatial position with varying temporal positions, as shown in Figure~\ref{fig:math_goal}. In this example, we assume tokens $x_2$, $x_3$ and $x_5$ are pruned and our goal is to find a function which operates on $\mathcal{P}(x_t) $ and satisfies Eqn. \ref{eqn:msa_equivalence}. Note that the derivation for this example extends naturally to the general case.

To be compatible with FlashAttention~\citep{dao2022flashattention}, the proposed function must operate outside the core attention calculation. Specifically, it is restricted to modifying either the input vectors ($q$, $k$, $v$) before the attention calculation or the attention output afterward. Mathematically, our goal is to define functions $f$ and $g$ such that the attention output computed from the kept tokens approximates the original output:
\begin{equation}
\begin{split}
    \frac{\sum_{j \in \mathcal{R}} g(e^{q^T f(k_j, c_j)}) v_j}{\sum_{j \in \mathcal{R}} g(e^{q^T f(k_j, c_j)})} \approx \frac{\sum_{i=1}^N e^{q^T k_i} v_i}{\sum_{i=1}^N e^{q^T k_i}}
\end{split}
\label{eqn:expansion}
\end{equation}
where $N$ is the total number of (unpruned) tokens, $\mathcal{R}$ denotes the set of indices for the tokens that remain after pruning, and $c_j$ represents the number of tokens approximated by the unpruned (remaining) token $j$  (such that $\sum_{j \in \mathcal{R}} c_j = N$).  We require that the approximation error is bounded by $O(\delta)$, where $\delta$ represents the maximum token approximation error (defined below).

Pruning temporally unchanged tokens ensures that the underlying tokens have similar values, \textit{i.e.}, $k_1 \approx k_2 \approx k_3$ and $k_4 \approx k_5$, as shown in Figure \ref{fig:math_goal}. (Note that, in this approximation, we disregard the impact of different noise values added to each token, which will be addressed in Section \ref{sec:iid_noise}). Furthermore, RoPE~\citep{su2021roformer} introduces position-dependent variations in attention, requiring explicit handling of rotational effects.

Replacing the keys from the pruning approximation into the original MSA calculation yields an expanded form of the attention output computed over the full sequence:
\begin{equation}
\label{eqn:rope_expansion}
\begin{aligned}
    \frac{\sum_{i=1}^N e^{q^T k_i} v_i}{\sum_{i=1}^N e^{q^T k_i}}
    \approx \frac{
        \sum_{j \in \mathcal{R}} \big( \sum_{m=0}^{c_j-1} e^{q^T (e^{m \theta \mathbf{i}} k_j)} \big) v_j
    }{
        \sum_{j \in \mathcal{R}} \big( \sum_{m=0}^{c_j-1} e^{q^T (e^{m \theta \mathbf{i}} k_j)} \big)
    }
\end{aligned}
\end{equation}
where $c_j$ is the number of tokens approximated by the kept token $j$ and $m\theta$ is the angle induced by RoPE. Note that the approximation error is bounded by $O(\delta)$, where $\delta = \max_{i,j,m} \|k_i - e^{m \theta \mathbf{i}} k_j\|$, due to the Lipschitz continuity of the self-attention calculation with respect to the keys. Consequently, combining Eqn. \ref{eqn:expansion} and Eqn. \ref{eqn:rope_expansion}, the objective further simplifies to finding $f$ and $g$ such that, for any query $q$ from the remaining tokens, the following approximation holds:
\begin{equation}
  g(e^{q^T f(k_j, c_j)}) \approx \sum_{m=0}^{c_j-1} e^{q^T (e^{m \theta \mathbf{i}} k_j)} 
\end{equation}
\subsection{The Impact of I.I.D. Noise}
\label{sec:iid_noise}
Although two temporally redundant patches may have similar values ($P_1 \approx P_2$) in the latent space, each is perturbed by independent Gaussian noise $\epsilon_i \sim \mathcal{N}(0, I)$. Naively assuming the resulting tokens are close ($x_1 \approx x_2$) ignores this independence, leading to visual artifacts, as shown below.



Suppose we decompose each token into a clean component and noise, \textit{i.e.}, $x_i = \bar{x}_i + \epsilon_i$, where the bar notation ($\bar{\cdot}$) denotes the noise-free (signal) component, and $W_Q$, $W_K$, $W_V$ are the projection weight matrices. The query, key, and value vectors are then:
\begin{equation}
q_i = \bar{q}_i + W_Q \epsilon_i, \text{ }
k_i = \bar{k}_i + W_K \epsilon_i, \text{ }
v_i = \bar{v}_i + W_V \epsilon_i
\label{eqn:projections}
\end{equation}
For a fixed query, the attention output over $N$ tokens is $\sum_{i=1}^N \sigma\left( \frac{q^T k_i}{\sqrt{D}} \right) v_j$, where $\sigma(\cdot)$ denotes softmax and $D$ is the token dimension. Expanding the dot product gives:
\begin{equation}
q^T k_i = \bar{q}^T \bar{k_i} + \bar{q}^T W_K \epsilon_i + \epsilon^T W_Q^T \bar{k_i} + \epsilon^T W_Q^T W_K \epsilon_i
\label{eqn:expanded}
\end{equation}
where $\epsilon$ and $\epsilon_i$ are the noise added to $q$ and $k_i$ respectively. Let $W_Q^T W_K \approx I$ for illustrative purposes (for the general case where there is no restriction on $W_Q^T W_K$, please refer to Appendix \ref{sec:noise_impact}, where our conclusions still hold). Assume $x_1\approx x_2$ means $\epsilon \approx \epsilon_i$. 

This leads to two critical results: 
\textbf{1. Attention Score Calculation:} The quadratic noise term $\epsilon_i^T \epsilon_j$ changes distribution from the Gaussian Distribution $\mathcal{N}$ to Chi squared $\chi^2$ distribution.
\begin{equation}
\epsilon_i^T \epsilon_j \;\sim\;
\begin{cases}
\mathcal{N}(0,D)  & \text{if } \epsilon_i \neq \epsilon_j \quad \text{(independent)} \\
\chi^2_D           & \text{if } \epsilon_i = \epsilon_j \quad \text{(duplicated)}
\end{cases}
\label{eqn:noise_interact}
\end{equation}
Note that by Central Limit Theorem, $\mathcal{N}(0,D)$ is an approximation for large token dimension $D$. The duplicated case introduces a large positive bias ($\mathbb{E}[\chi^2_D] = D$) and higher variance ($2D$), inflating attention weights on duplicated tokens.

\textbf{2. Value Aggregation:} Duplication changes the summed noise from

\noindent   $W_V \sum_{j=1}^n \epsilon_j$ ( variance $O(nI_D)$) to $n W_V \epsilon$ (variance $O(n^2I_D)$), resulting in quadratic variance explosion.

Empirically, forcing $ x_1 = x_2$ by duplication produces strong, noisy patterns and significantly degrades the quality of the generated videos, as shown in Section \ref{sec:noise_duplication}, highlighting the importance of accounting for I.I.D. noise.

\section{Methods}
\begin{figure*}[htp]
  \centering
  \vspace{-4em}
  \includegraphics[width=0.9\linewidth]{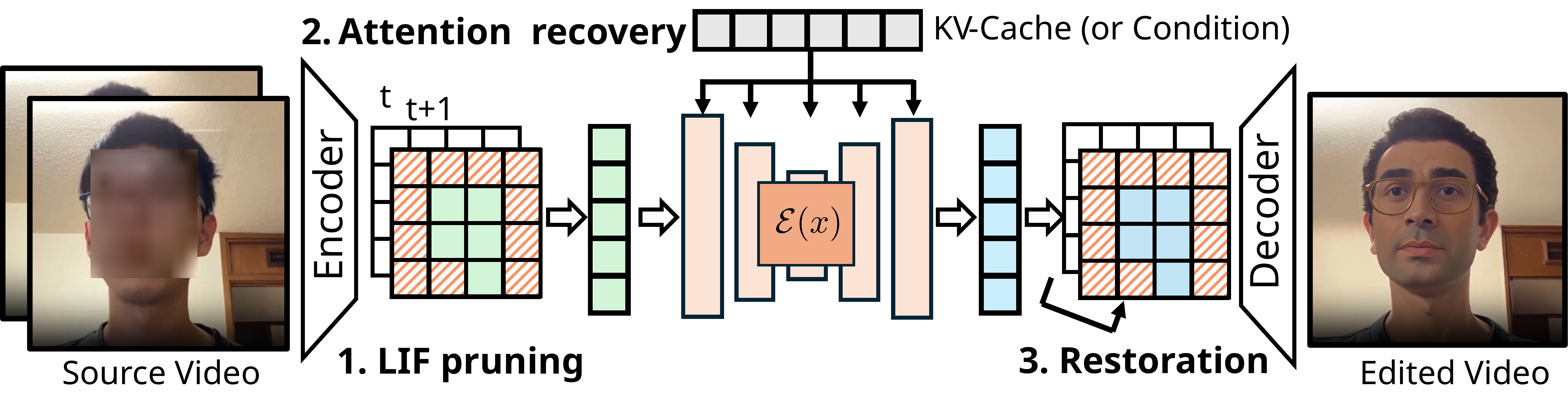}
  \caption{LIPAR consists of three stages: 1. Pruning 2. Attention Recovery and 3. Restoration.}
  \label{fig:framework}
\end{figure*}

\begin{wrapfigure}{r}{0.5\linewidth}
  \centering
  \vspace{-5em}
  \includegraphics[width=1\linewidth]{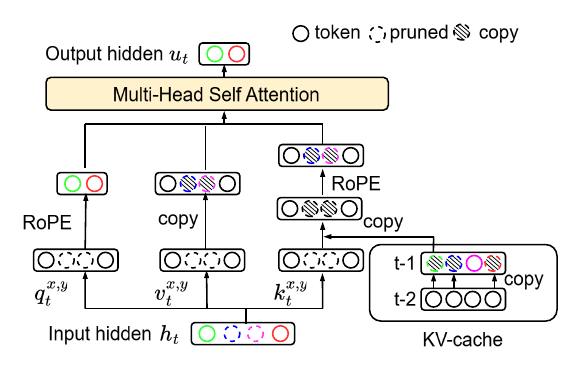}
  \caption{\textbf{Attention Recovery.} Pruned keys ($k$) and values ($v$) are approximated by copying their temporal counterparts from the clean KV-cache.}
  
  \vspace{-2em}
  \label{fig:attention_recovery}
\end{wrapfigure}


\subsection{LIPAR Overview}
\label{sec:LIPAR_overview}
In Figure~\ref{fig:framework}, we present an overview of the proposed pruning framework, which operates in three stages to accelerate conditioned video generation. First, \textbf{Latent Inter-Frame Pruning} removes temporally redundant patches in the latent space. However, directly removing tokens induces a training-inference discrepancy, leading to visual artifacts. To mitigate this, we propose \textbf{Attention Recovery}. Finally, the \textbf{Restoration} step upsamples the token count for decoding and maps the latents back to pixel space.


\subsection{Token Pruning and Restoration} 
\label{sec:token_pruning}
\paragraph{Latent Inter-Frame Pruning.}
The diffusion latent space contains temporal redundancy. We propose Latent Inter-frame Pruning to identify and skip computation on unchanged patches by comparing temporally consecutive patches at the same spatial location: $\|p_{t}^{x,y} - p_{t+1}^{x,y}\|_1 < \tau$, where $\tau$ is a predefined threshold determining whether the temporal difference is small enough to consider the patch unchanged. Due to the high compression rate of the latent space, subtle movements can yield differences that fall below this threshold, leading to the erroneous pruning of dynamic tokens. During restoration, erroneously reusing these tokens repeats the subtle motions, which manifest as glitches upon decoding and degrade video quality. To detect such subtle movements, we integrate \textbf{motion detection for frame difference} into LIF pruning by leveraging spatial and temporal information from neighboring tokens, capturing object-level dynamics rather than isolated pixel changes. See Appendix Alg.~\ref{alg:LIF_pruning} for the full algorithm.

\paragraph{Latent Patch Restoration.}
After the denoising process, the Diffusion Transformer outputs a set of pruned and denoised patches. However, since decoding requires patches with fixed dimensions, we must restore them. To achieve this, we reconstruct the pruned patches by duplicating the corresponding patches from the previous frame. Alg.~\ref{alg:unprune} in Appendix~\ref{apx:algorithms} details this restoration procedure.

\subsection{Attention Recovery}
\label{sec:attention_recovery}
Figure~\ref{fig:attention_recovery} illustrates the Attention Recovery mechanism applied to the causal attention backbone~\citep{yin2025causvid, huang2025selfforcing}. The method relies on two core mechanisms: M-Degree Approximation and Noise-Aware Duplication. M-Degree Approximation ensures that the $m$ closest keys and values to the query remain unpruned by copying Key (K) and Value (V) vectors from their temporal counterparts. Noise-Aware Duplication restricts copying to ``clean'' tokens, \textit{i.e.}, those from the KV cache. The method also extends to bidirectional attention, as demonstrated in Section~\ref{sec:ttm}.


\paragraph{M-degree Approximation.}
To recover self-attention values from the pruned sequence, we seek functions $f$ and $g$ approximating the exponential sum in Eqn.~\ref{eqn:objective} (as discussed in Section~\ref{sec:proble_formulation}), with approximation error bounded by $O(\delta)$, where $\delta$ is the maximum token approximation error. Here, $e^{\theta \mathbf{i}}$ is the RoPE rotation matrix, $q$ is the query after RoPE, and $c_j$ is the number of tokens approximated by kept token $j$. The middle term is the full attention sum we wish to approximate. By a \textbf{log-sum-exp} (LSE) argument, it is tightly lower-bounded by its $m$ largest terms (right term):
\begin{equation}
  g(e^{q^T f(k_j, c_j)}) \approx \sum_{l=0}^{c_j-1} e^{q^T (e^{l \theta \mathbf{i}} k_j)} \approx \sum_{l \in \mathcal{M}} e^{q^T (e^{l \theta \mathbf{i}} k_j)}
  \label{eqn:objective}
\end{equation}
where $\mathcal{M}$ indexes the $m$ largest exponents. Maximizing the exponent is equivalent to minimizing the angular deviation $|l\theta - \phi|$ between $q$ and the rotated keys, where $\phi$ is the query's rotation angle. Since queries in causal attention correspond to the most recent tokens, $\phi$ aligns best with the latest keys; thus $\mathcal{M}$ is simply the $m$ most recent indices, giving:
\begin{equation}
f(k_j, c_j) = \left( e^{l \theta \mathbf{i}} k_j \right)_{l=0}^{c_j-1}, \quad  g(X) = \sum_{l=c_j-m}^{c_j-1} X_l.
\label{eqn:func_f}
\end{equation}
Even at full duplication ($m=N$), we retain a linear speedup: fewer queries in self-attention means fewer tokens flow through \textbf{all} subsequent layers (FFN, cross-attention), yielding a $\frac{N_{\text{total}}}{N_{\text{kept}}}$ speedup. LIPAR is also compatible with FlashAttention~\citep{dao2022flashattention}.  The $m$-degree approximation enables the pruning of redundant tokens, thus reducing the GPU memory usage and computational complexity in attention layers.

\paragraph{Noise-Aware Duplication.}
\label{sec:noise_aware_duplication}
Although Equation~\ref{eqn:func_f} suggests a straightforward solution to Attention Recovery, this method may fail in practice, as discussed in Section~\ref{sec:iid_noise}. To address this, we propose \emph{Noise-Aware Duplication}, which duplicates only the clean component of tokens to prevent the noise correlations during the self-attention computation.

We achieve this by duplicating the temporally closest clean tokens from the KV cache. These tokens are inherently clean as they are produced by the model's own final denoising step at zero noise level as part of standard video generation. However, this introduces a new challenge: while the previous approximation assumed $X_{t-1} \approx X_t$ for pruned $X_t$, we now approximate $X_t$ using $X_{t-k}$, where $k$ is the temporal offset and $X_{t-k}$ is the closest clean token in the KV cache. To ensure a valid approximation, we add a long-term difference constraint to LIPAR: a token is pruned only if \emph{both} short- and long-term differences are satisfied, where $\tau_2$ is the threshold for the long-term difference and $S$ is the denoising block size:
\begin{equation}
    \lVert X_{t-k} - X_t \rVert_1 < \tau_2, \quad k = 
    \begin{cases} 
        1 & \text{if } t \equiv 0 \pmod{S}, \\
        t - S \lfloor t/S \rfloor & \text{otherwise}.
    \end{cases}
    \label{eqn:second_mask}
\end{equation}

\section{Experiments}
\subsection{Experimental Setup}
We implement LIPAR on top of Self-Forcing~\citep{huang2025selfforcing}, and follow CausVid~\citep{yin2025causvid} and StreamV2V~\citep{liang2024looking} in using SDEdit~\citep{meng2022sdedit} for video-to-video translation. By default, we use a 4-step denoising schedule with initial noise level $t=400$ (out of $1000$). The pruning thresholds $\tau_1$ and $\tau_2$ (in Section~\ref{sec:token_pruning} and Eq.~\ref{eqn:second_mask}) are set to $0.15$ and $0.3$, respectively. Following prior works~\citep{tokenflow2023, liang2024looking}, we evaluate at $480 \times 832$ resolution on $102$ video-prompt pairs from DAVIS 2017 train-val~\citep{davis2017dataset} and $120$ from OpenHumanVid~\citep{li2024openhumanvid}, covering diverse objects (\textit{e.g.}, humans, animals, cars) and human portraits, respectively and use  Warp Error~\citep{Lai-ECCV-2018}, CLIP Score~\citep{radford21clip}, and VBench~\citep{huang2023vbench} as evaluation metrics. All experiments run on an NVIDIA A6000 GPU unless otherwise specified. See Appendix~\ref{appendix:experimental_settings} for full settings.

\begin{figure*}[!ht]
    \centering
    \vspace{-6em}
    \includegraphics[width=0.95\linewidth]{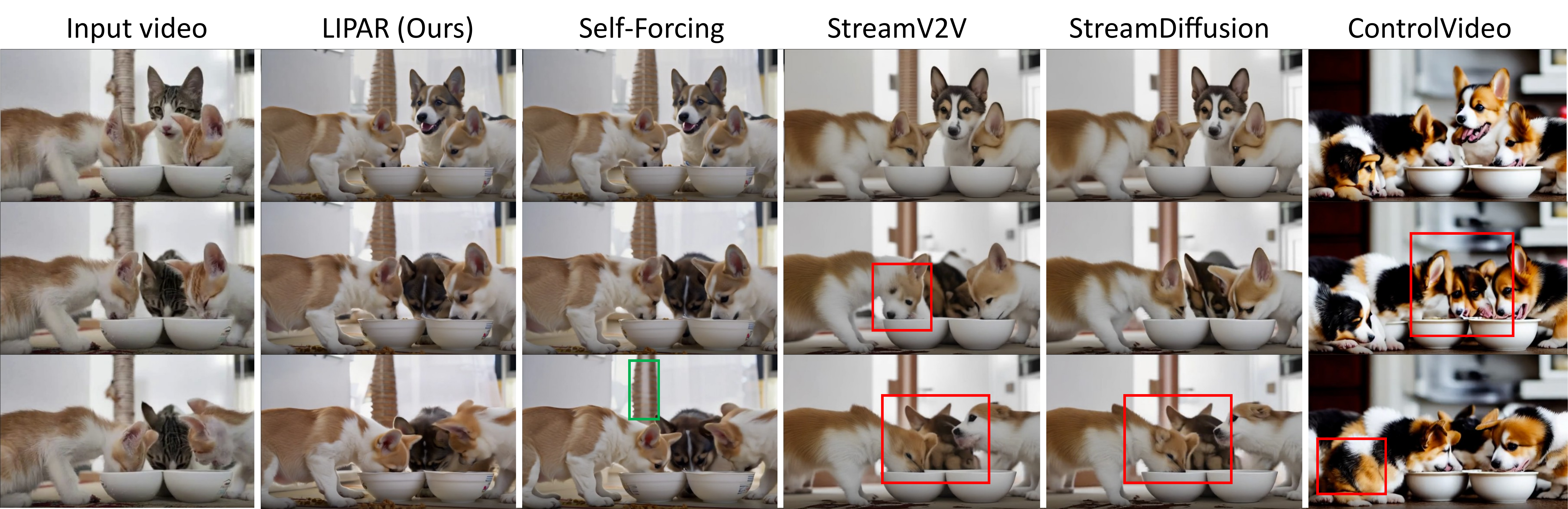}
    \caption{\textbf{Qualitative comparison with representative low latency V2V models.} Prompt: \textit{``Three corgi puppies sharing a meal together on a kitchen floor.''} }
    \label{fig:vis_diff_models}

\end{figure*}

\subsection{Comparison with Real-time Video Editing Models}
\paragraph{Qualitative Comparisons.}
In Figure~\ref{fig:vis_diff_models}, we qualitatively compare LIPAR against SOTA V2V models. StreamDiffusion and StreamV2V suffer from flickering and structural defects, with the dogs' heads merging into unnatural shapes (red rectangles). ControlVideo~\citep{zhang2023controlvideo} also suffers from structural defects (fused dog faces) despite having stronger editing effects. Self-Forcing generates high-quality videos, but re-editing temporally unchanged tokens incurs unnecessary cost and introduces temporal instability, causing subtle background fluctuations (green rectangle). In contrast, LIPAR inherits the high generation quality of Self-Forcing, producing dogs that mimic real ones, while avoiding the re-editing of unchanged tokens, yielding a stable background. We encourage readers to view the supplementary materials for more examples.
\paragraph{Quantitative Comparisons.}
As shown in Table~\ref{tab:dataset_comparison}, LIPAR matches or outperforms all baselines on every metric, except Subject and Background Consistency on OpenHumanVid, where ControlVideo scores higher. This is due to ControlVideo's limited motion, as evidenced by its higher warp error and lower dynamic degree ($0.18$ compared to LIPAR's $0.50$). The gap between LIPAR and Self-Forcing is small, reflected in the near-identical edited results between the two, but LIPAR is significantly more compute-efficient. As shown in Figure~\ref{fig:eval}, LIPAR achieves the highest FPS among all methods, being $1.45\times$ and $1.57\times$ faster than Self-Forcing on DAVIS and OpenHumanVid, respectively.

\begin{table}[t]
    \vspace{0.3em}
    \centering
    \begin{minipage}[c]{0.57\textwidth}
        \centering
        \renewcommand{\arraystretch}{0.92}
        \resizebox{\linewidth}{!}{%
        \begin{tabular}{lccccccc}
            \toprule
             & & & & \multicolumn{4}{c}{\textbf{V-Bench Quality} $\uparrow$} \\
            \cmidrule(lr){5-8}
            \textbf{Method} & \textbf{FPS} $\uparrow$ & \textbf{Warp Err.} $\downarrow$ & \textbf{CLIP} $\uparrow$ & \textbf{Subj.} & \textbf{Backg.} & \textbf{Motion} & \textbf{Img. Qual.} \\
            \midrule
            \multicolumn{8}{l}{\textit{DAVIS}} \\
            StreamDiffusion       & 9.1           & 87.4          & 0.955          & 0.888          & 0.911          & 0.981          & 0.668          \\
            StreamV2V             & 5.9           & 74.7          & 0.972          & 0.913          & 0.927          & 0.987          & 0.656          \\
            ControlVideo          & 0.19          & 102.1         & 0.976          & 0.921          & 0.938          & 0.979          & 0.594          \\
            Self-Forcing          & 8.4           & 70.6          & 0.980          & 0.922          & 0.936          & 0.988          & \textbf{0.676} \\
            \textbf{LIPAR (Ours)} & \textbf{12.2} & \textbf{58.6} & \textbf{0.981} & \textbf{0.925} & \textbf{0.939} & \textbf{0.989} & \textbf{0.676} \\
            \midrule
            \multicolumn{8}{l}{\textit{OpenHumanVid}} \\
            StreamDiffusion       & 9.2           & 92.4          & 0.962          & 0.910          & 0.905          & 0.986          & 0.671          \\
            StreamV2V             & 4.8           & 77.4          & 0.981          & 0.934          & 0.922          & 0.991          & 0.670          \\
            ControlVideo          & 0.12          & 102.7         & 0.987          & \textbf{0.951} & \textbf{0.956} & 0.981          & 0.675          \\
            Self-Forcing          & 8.1           & 78.3          & 0.989          & 0.944          & 0.947          & 0.992          & \textbf{0.688} \\
            \textbf{LIPAR (Ours)} & \textbf{12.7} & \textbf{71.5} & \textbf{0.990} & 0.948          & 0.952          & \textbf{0.993} & \textbf{0.688} \\
            \bottomrule
        \end{tabular}}
        \vspace{3pt}
        \caption{Results for real-time video editing.}
        \label{tab:dataset_comparison}
    \end{minipage}\hfill
    \begin{minipage}[c]{0.38\textwidth}
    \vspace{-.7 em}
        \centering
        \includegraphics[width=\linewidth]{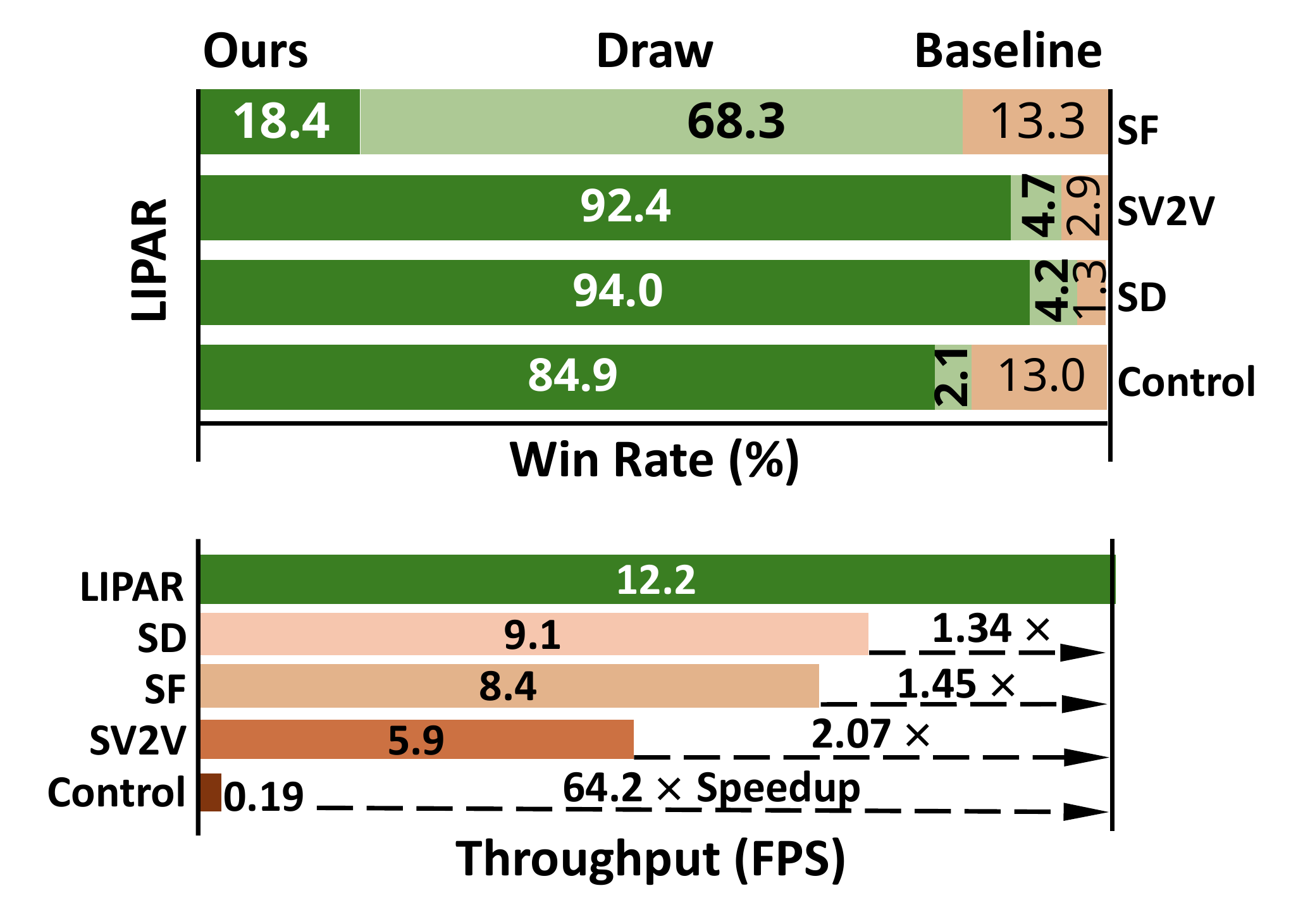}
        \captionof{figure}{Human eval and throughput}
        \label{fig:eval}
    \end{minipage}
    \vspace{-12pt}
\end{table}
\paragraph{Human Evaluation.}
We also conduct a human evaluation with $16$ participants, each performing at least $100$ pairwise comparisons on the DAVIS dataset, chosen for its wide range of objects. Following TokenFlow~\citep{tokenflow2023} and StreamV2V \citep{liang2024looking}, we use a Two-Alternative Forced Choice protocol, where participants select the better of two side-by-side videos generated from the same video-prompt pair from different models. See Appendix~\ref{appendix:human_eval_test} for the evaluation webpage. 

In Figure~\ref{fig:eval}, participants reported no noticeable difference between LIPAR and Self-Forcing (SF) in $68.3\%$ of cases. Furthermore, in $18.4\%$ of cases, participants preferred LIPAR, likely due to its more consistent background generation. LIPAR demonstrates a clear advantage over other real-time competitors, achieving win rates above $84\%$ against StreamDiffusion (SD), StreamV2V (SV2V), and ControlVideo (Control). These results confirm that our method significantly outperforms prior state-of-the-art low-latency models while delivering higher throughput.

\subsection{Online Video Editing for Long-Horizon Stability}
  \label{sec:long_horizon}
To demonstrate LIPAR's streaming capability, we integrate it with a live camera for online video editing. As source frames arrive, the pruning mask is computed once per frame using only current and past source latents, preserving the streaming capability of the causal backbone. We provide the streaming demo in the supplementary material.

Using this streaming architecture, we evaluate LIPAR on 12 one-minute video-prompt pairs from DH-FaceVid-1K~\citep{di2025facevid}. To mitigate error accumulation from repeated token reuse, we introduce a periodic refresh inspired by the I-/P-frame structure of MPEG~\citep{MPEG1991}: every six latent frames, pruning is disabled in the final denoising step (1 of 4), adding approximately $1.3\%$ of the unpruned computation. We measure VBench quality scores~\citep{huang2023vbench} over non-overlapping five-second windows. As shown in Table~\ref{tab:long_horizon}, quality remains stable throughout the sequence. A per-video linear fit gives a change of $+0.0002$ per ten seconds ($95\%$ CI: $[-0.0007,+0.0010]$), indicating no measurable quality decline over the evaluated duration. See Appendix~\ref{app:long_horizon_results} for the full results.

\begin{table}[t]
      \vspace{-0.5em}
      \centering
      \footnotesize
      \setlength{\tabcolsep}{5pt}
      \caption{Long-horizon results with periodic refresh
      (12 video-prompt pairs). Time is the start of each
      five-second window; the mean covers all twelve windows.}
      \label{tab:long_horizon}
      \begin{tabular}{lccccccc}
          \toprule
          \textbf{Time (s)} & \textbf{Quality} $\uparrow$ &
          \textbf{Subj.} $\uparrow$ & \textbf{Backg.} $
          \uparrow$ & \textbf{Flicker} $\uparrow$ &
          \textbf{Motion} $\uparrow$ & \textbf{Aesth.} $
          \uparrow$ & \textbf{Img.} $\uparrow$ \\
          \midrule
          0    & 0.856 & 0.959 & 0.954 & 0.990 & 0.994 &
          0.579 & 0.709 \\
          30   & 0.856 & 0.964 & 0.957 & 0.989 & 0.994 &
          0.575 & 0.705 \\
          55   & 0.857 & 0.970 & 0.961 & 0.990 & 0.993 &
          0.569 & 0.704 \\
          \midrule
          Mean & 0.855 & 0.962 & 0.956 & 0.989 & 0.994 &
          0.570 & 0.708 \\
          \bottomrule
      \end{tabular}
  \end{table}

\subsection{Comparison with Pruning Baselines}

  \begin{wraptable}{r}{0.48\textwidth}
      \vspace{-1em}
      \centering
      \scriptsize
      \setlength{\tabcolsep}{2pt}
      \caption{Comparison with pruning baselines.}
      \label{tab:pruning_comparison}
      \resizebox{\linewidth}{!}{%
      \begin{tabular}{lccccccc}
        \toprule
         & & & & \multicolumn{4}{c}{\textbf{V-Bench Quality} $\uparrow$} \\
        \cmidrule(lr){5-8}
        \textbf{Method} & \textbf{Rate} & \textbf{FPS} $\uparrow$ & \textbf{Warp} $\downarrow$ & \textbf{Subj.} & \textbf{Backg.} & \textbf{Motion} & \textbf{Img.} \\
        \midrule
        Original              & 0    & 8.4           & 70.6          & 0.921          & 0.941          & 0.988          & 0.678          \\
        \midrule
        Important             & 0.32 & 9.2           & 83.4          & 0.852          & 0.917          & 0.988          & 0.577          \\
        IDM                   & 0.32 & 9.1           & 84.8          & 0.843          & 0.907          & 0.986          & 0.585          \\
        ToMe                  & 0.32 & 9.1           & 85.0          & 0.856          & 0.915          & 0.987          & 0.622          \\
        \textbf{LIPAR}        & 0.32 & \textbf{12.2} & \textbf{58.6} & \textbf{0.923} & \textbf{0.941} & \textbf{0.989} & \textbf{0.676} \\
        \midrule
        Important             & 0.20 & 8.5           & 77.4          & 0.887          & 0.924          & 0.988          & 0.633          \\
        IDM                   & 0.20 & 8.4           & 79.8          & 0.876          & 0.917          & 0.987          & 0.629          \\
        ToMe                  & 0.20 & 8.4           & 80.4          & 0.883          & 0.928          & 0.988          & 0.653          \\
        \textbf{LIPAR}        & 0.20 & \textbf{10.9} & \textbf{63.5} & \textbf{0.921} & \textbf{0.940} & \textbf{0.989} & \textbf{0.676} \\
        \midrule
        Important             & 0.10 & 8.3           & 74.3          & 0.909          & 0.930          & 0.988          & 0.661          \\
        IDM                   & 0.10 & 8.2           & 75.0          & 0.903          & 0.930          & 0.988          & 0.653          \\
        ToMe                  & 0.10 & 8.2           & 75.9          & 0.903          & 0.930          & 0.988          & 0.668          \\
        \textbf{LIPAR}        & 0.10 & \textbf{9.8}  & \textbf{68.3} & \textbf{0.920} & \textbf{0.940} & 0.988          & \textbf{0.677} \\
         \bottomrule
      \end{tabular}%
      }
      \vspace{-1em}
  \end{wraptable}

We compare LIPAR against state-of-the-art training-free pruning methods, including ToMe~\citep{bolya2023tomesd}, Importance-based Token Merging~\citep{wu2025importancetome}, and IDM~\citep{fang2025attend}. Following ToMe's official implementation, we integrate these methods into Self-Forcing, restricting merging operations to the Self-Attention layers and unmerging tokens before the Cross-Attention layers. We fix the pruning rates across all methods and compare them at three rates: 10\%, 20\%, and 32\%.

\begin{figure}[t]
    \centering
    \includegraphics[width=0.9\linewidth]{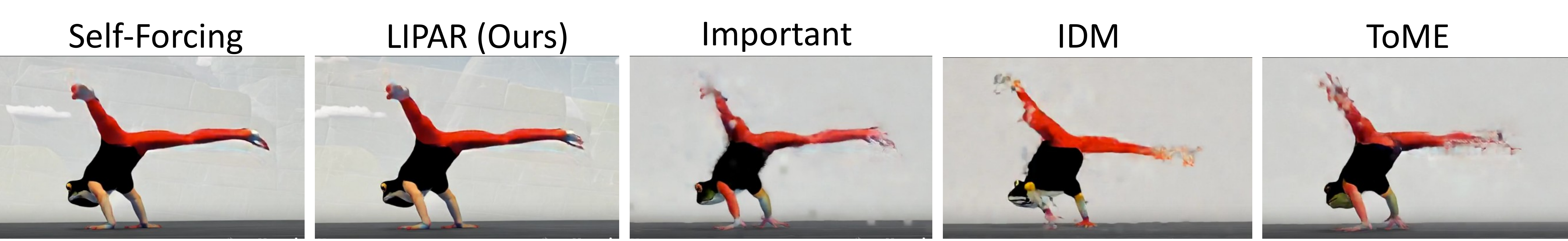}
    \vspace{-0.5em}
    \caption{\textbf{Visual comparison of different pruning methods.} LIPAR achieves better visual quality. Prompt: \textit{``Animation style of a frog dancing and performing acrobatic side somersaults.''}}
    \label{fig:diff_prune_vis}
    \vspace{-0.8em}
\end{figure}

Figure~\ref{fig:diff_prune_vis} qualitatively compares LIPAR against training-free pruning methods. Video from LIPAR closely preserves the fidelity of the original Self-Forcing model. In contrast, IDM and ToMe suffer from severe blurring on the frog's body, and Importance-based Token Merging introduces noticeable artifacts with inconsistent coloration. LIPAR demonstrates a clear advantage over the rest. Table~\ref{tab:pruning_comparison} quantitatively evaluates the generated videos. Consistent with the visualization, LIPAR outperforms all pruning methods across nearly all metrics. This performance gap becomes more pronounced as the pruning rate increases. Additionally, LIPAR achieves a faster FPS due to its end-to-end pruning and cheap pruning operations. See Appendix~\ref{appendix:compare pruning methods} for more observations.

\vspace{-0.4em}
\subsection{Motion-Controlled Video Generation}
\label{sec:ttm}
We apply LIPAR on Time-to-Move (TTM)~\citep{singer2025timetomove} to demonstrate \textbf{generalizability on a new task and modality} (\textit{i.e.}, users drag segments on an image as the motion trajectory for video generation) and \textbf{on a new model} (\textit{i.e.}, bi-directional and MoE diffusion model WAN2.2~\citep{wan2025}). As shown in Table~\ref{tab:ttm_comparison} and Figure~\ref{fig:combined_ablation} (a), LIPAR maintains performance comparable to the baseline TTM while achieving a $1.5\times$ increase in denoising throughput, confirming the generalizability of LIPAR.



\begin{table}[t]
    \vspace{-0.8em}
    \centering
    \captionsetup{font=footnotesize, skip=3pt}
    \begin{minipage}[c]{0.50\textwidth}
        \centering
        \scriptsize
        \setlength{\tabcolsep}{3.5pt}
        \renewcommand{\arraystretch}{1.0}
        \caption{Quantitative comparison on TTM.}
        \label{tab:ttm_comparison}
                \begin{tabular}{lcccccc}
            \toprule
            & & & \multicolumn{4}{c}{\textbf{V-Bench} $\uparrow$} \\
            \cmidrule(lr){4-7}
            \textbf{Method} & \textbf{FPS}$\uparrow$ & \textbf{Warp}$\downarrow$ & \textbf{Subj.} & \textbf{Backg.} & \textbf{Mot.} & \textbf{Img.} \\
            \midrule
            TTM (5b) & 0.58 & 63.6 & 0.956 & 0.957 & 0.991 & 0.652 \\
            \textbf{LIPAR} & \textbf{0.87} & \textbf{40.4} & \textbf{0.961} & \textbf{0.962} & \textbf{0.993} & \textbf{0.665} \\
            \bottomrule
        \end{tabular}

        \vspace{8pt}
        \setlength{\tabcolsep}{4pt}
        \caption{Threshold sensitivity on DAVIS (defaults in \textbf{bold}).}
        \label{tab:threshold_sensitivity}
                \begin{tabular}{ccc@{\hspace{1.2em}}ccc}
            \toprule
            \multicolumn{3}{c}{Varying $\tau_1$ ($\tau_2=0.3$)} & \multicolumn{3}{c}{Varying $\tau_2$ ($\tau_1=0.15$)} \\
            \cmidrule(lr){1-3}
            \cmidrule(lr){4-6}
            $\tau_1$ & \textbf{Warp} $\downarrow$ & \textbf{Quality} $\uparrow$ & $\tau_2$ & \textbf{Warp} $\downarrow$ & \textbf{Quality} $\uparrow$ \\
            \midrule
            0.11 & 62.83 & 0.832 & 0.15 & 60.74 & 0.833 \\
            0.13 & 59.62 & 0.833 & 0.20 & 59.11 & 0.833 \\
            \textbf{0.15} & \textbf{58.58} & \textbf{0.834} & 0.25 & 58.61 & 0.834 \\
            0.17 & 58.78 & 0.833 & \textbf{0.30} & \textbf{58.58} & \textbf{0.834} \\
            0.19 & 60.19 & 0.832 & 0.35 & 58.58 & 0.834 \\
            0.21 & 64.26 & 0.831 & 0.40 & 58.55 & 0.834 \\
            0.23 & 68.16 & 0.827 & 0.45 & 58.61 & 0.834 \\
            \bottomrule
        \end{tabular}

        \vspace{8pt}
        \setlength{\tabcolsep}{4pt}
        \caption{Streaming performance on an NVIDIA A6000.}
        \label{tab:streaming_metrics}
                \begin{tabular}{lcccc}
            \toprule
            & \multicolumn{2}{c}{DAVIS} & \multicolumn{2}{c}{OpenHumanVid} \\
            \cmidrule(lr){2-3}
            \cmidrule(lr){4-5}
            \textbf{Metric} & \textbf{LIPAR} & \textbf{SF} & \textbf{LIPAR} & \textbf{SF} \\
            \midrule
            First-frame lat. (ms) & 647  & 1000  & 673  & 1004 \\
            Per-frame time (ms) & 82.0 & 119.0 & 78.4 & 123.4 \\
            Mask time (ms) & 0.65 & --    & 0.69 & -- \\
            Peak GPU memory (GB)      & 18.6 & 26.2  & 18.6 & 26.2 \\
            \bottomrule
        \end{tabular}
        \vspace{12pt}
    \end{minipage}\hfill
    \begin{minipage}[c]{0.48\textwidth}
        \centering
        \includegraphics[width=\linewidth]{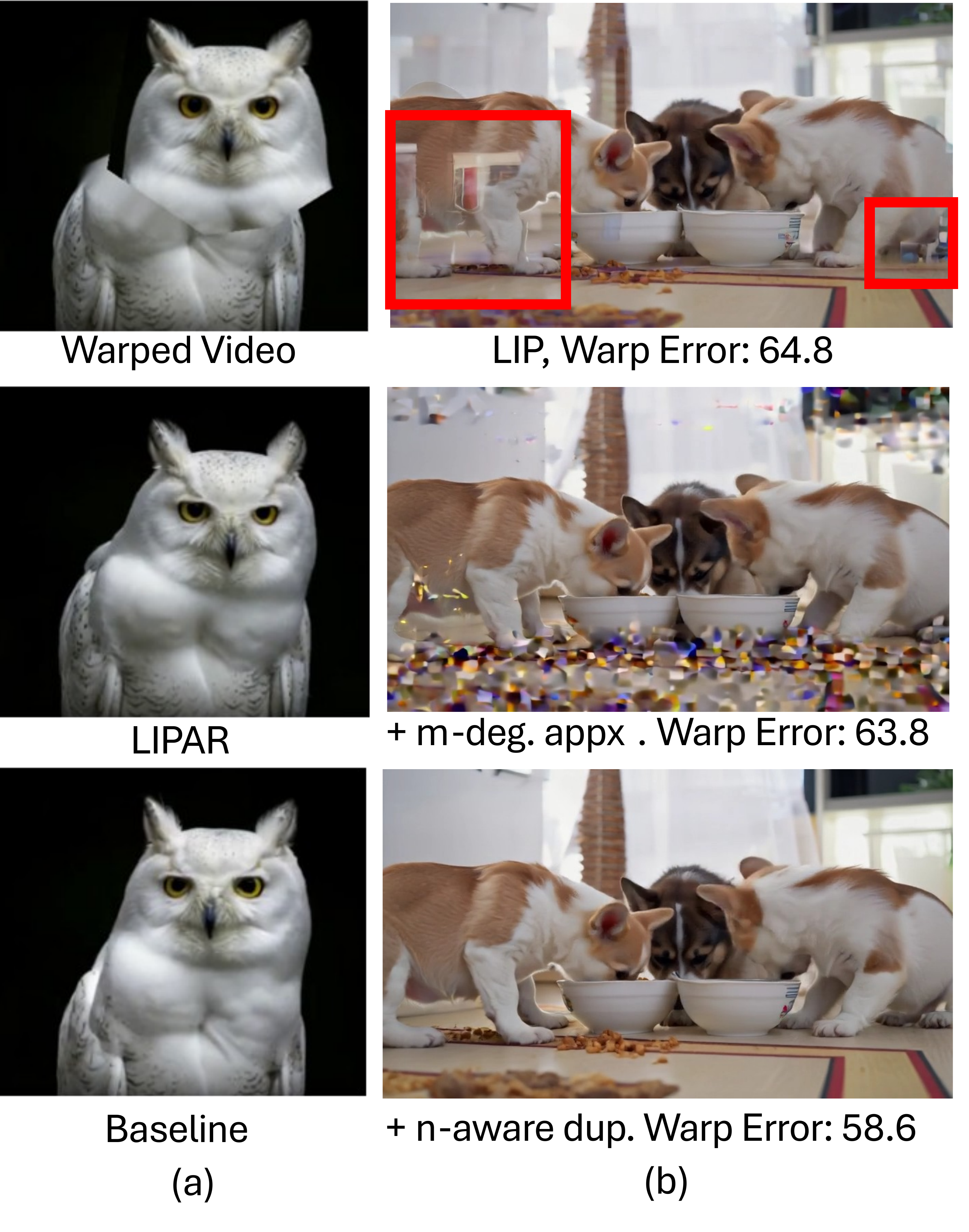}
        \vspace{-0.2em}
        \captionof{figure}{\textbf{(a)} Visual comparison on TTM. \textbf{(b)} Effectiveness of Attention Recovery.}
        \label{fig:combined_ablation}
    \end{minipage}
    \vspace{-1.2em}
\end{table}

\vspace{-0.5em}
\subsection{Ablation Study and Profiling}
\label{sec:noise_duplication}
\vspace{-0.3em}
\paragraph{Effectiveness of Attention Recovery.}
Figure~\ref{fig:combined_ablation}(b) demonstrates the effectiveness of Attention Recovery. Latent Inter-frame Pruning alone produces artifacts (red rectangles). Adding $m$-degree approximation introduces noisy artifacts by violating the diffusion model's i.i.d. noise assumption. Combining $m$-degree approximation with Noise-Aware Duplication, the complete Attention Recovery successfully preserves visual quality. Warp error measured \textbf{across the dataset}, reported in the figure, confirms these visual observations.

\vspace{-0.5em}
\paragraph{Sensitivity to Pruning Thresholds.}
We use the same thresholds, $\tau_1=0.15$ and $\tau_2=0.3$, across all 222 video-prompt pairs from DAVIS and OpenHumanVid without per-dataset adjustment. Table~\ref{tab:threshold_sensitivity} varies each threshold on DAVIS while fixing the other. Small $\tau_1$ retains most tokens, approaching the original Self-Forcing behavior, whereas large $\tau_1$ causes excessive pruning and degrades quality. Performance remains similar near the default $\tau_1=0.15$. For $\tau_2$, results approximately plateau from $0.25$ onward, supporting the selected default.

\vspace{-0.5em}
\paragraph{Runtime Profiling.}
We evaluate streaming performance on an NVIDIA A6000. As shown in Table~\ref{tab:streaming_metrics}, LIPAR reduces first-frame latency, average per-frame processing time, and peak GPU memory on both datasets. Mask computation takes only $0.65$\,ms per frame on DAVIS and $0.69$\,ms on OpenHumanVid, accounting for less than $1\%$ of the corresponding per-frame processing time. These results demonstrate that pruning adds little overhead while improving streaming efficiency.

We further modulate the pruning rate by adjusting the threshold $\tau$ in LIPAR and measure the corresponding latency. We observe a strong linear correlation (Pearson $r=0.999$) between the number of retained tokens and latency, verifying that LIPAR maintains $O(n)$ complexity even with Attention Recovery, where $n$ is the number of retained tokens. See Appendix~\ref{appdx:latency_token_number} for details.

\vspace{-1.0em}
\section{Conclusion and Future Work}
\vspace{-0.7em}
We identify a strong correlation between pixel-space and latent-space dynamics, enabling the adaptation of traditional video compression to Diffusion Transformers. We further formalize the condition that pruning must satisfy to preserve generation quality, and propose Attention Recovery to close the resulting train-inference gap. LIPAR achieves an average $1.45\times$ inference speedup with high visual fidelity, outperforming existing training-free pruning methods. We view this as a foundational step toward integrating pixel-level video compression with latent video generation. While LIPAR currently relies on a source video for the pruning mask, the gradual refinement of diffusion makes extension to T2V feasible for future work.






\begin{ack}
We thank Ruichen Jiang for insightful discussions regarding the mathematical derivations in the paper.

\paragraph{Funding and competing interests.} This work received no third-party funding. The authors declare no competing interests.
\end{ack}

\bibliographystyle{plainnat}   
\bibliography{references}      


\newpage

\appendix
\section{Broader Impact}
Our work aims to enable efficient and temporally consistent video synthesis for beneficial applications. As with other generative models, however, it may be misused for fraudulent or otherwise harmful purposes. We strongly oppose any use of this technology to produce misleading or harmful content. We also note that videos generated by our model retain subtle artifacts that remain identifiable by human observers.

\section{Latents Compression Experiment}
\label{appedix:latents_compression}
\begin{figure}[h]
    \centering
    \includegraphics[width=0.75\linewidth]{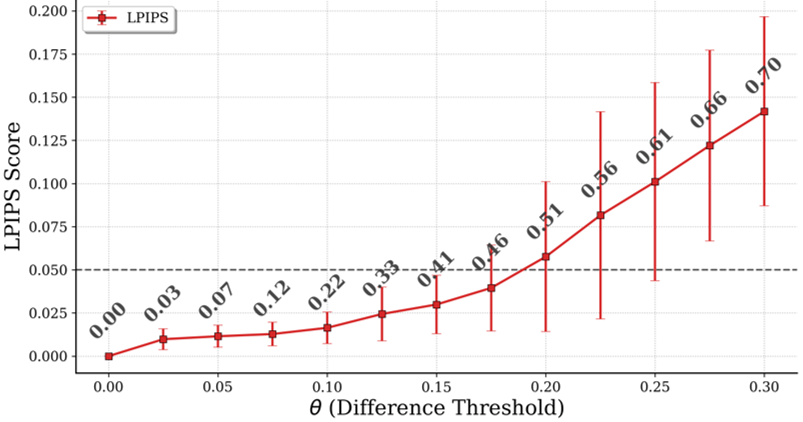}
    \caption{LPIPS Score\textit{ vs.} $\theta$. As we increase the threshold $\theta$ for compression in Eqn. \ref{eqn:vae_observation}, the compression rate (annotated in black) increases. Notably, high visual similarity (LPIPS $\le 0.05$, dashed line) is maintained even when the compression rate rises to $46\%$. This quantitatively confirms that substantial temporal redundancy exists in latent space.}
    \label{fig:lpips_tradeoff}
\end{figure}
There is no guarantee that the temporal redundancy exists in the latent space, despite the redundancy observed in the input video at pixel space. This is because the latent space is heavily compressed via the encoder, making it difficult to determine the real semantics of each latent patch. However, the existence of redundancy is central to pruning methods. As a result, we must verify this property in advance.

To measure the temporal redundancy in latent space, we conduct an experiment where patches $p_{t+1}^{x,y}$ are replaced (which we refer to as compressed) by their temporal predecessors (if sufficiently similar), and observe whether this influences the decoded results. Mathematically, we formulate this compression as:
\begin{equation}
    \hat{p}_{t+1}^{x,y} = 
    \begin{cases} 
        p_t^{x,y} & \text{if } \lVert p_{t+1}^{x,y} - p_t^{x,y} \rVert_1 < \theta \\
        p_{t+1}^{x,y} & \text{otherwise}
    \end{cases}
    \label{eqn:vae_observation}
\end{equation}
Here, $\hat{p}_{t+1}^{x,y}$ is the patch after compression, $\theta$ represents the threshold for judging similarity. To validate the fidelity of the compressed video, we require that the similarity between the compressed latents and the original decoded video exceeds a quality threshold $\tau$, \textit{i.e.}, 
$\text{Sim}\left( \text{Dec}(\hat{p}), \text{Dec}(p) \right) > \tau$, where $\text{Dec}(\cdot)$ denotes the decoder mapping latents to pixel space and $\text{Sim}(\cdot)$ is the similarity metric (we use LPIPS \citep{zhang2018perceptual} in this experiment). Figure~\ref{fig:compress_latents} shows an example where $44.5\%$ of the latent patches are compressed, yet the decoded results are similar to the uncompressed video, showing that temporal redundancy indeed exists.

To further validate these results, we select ten input videos and increase the compression rate by gradually increasing $\theta$, as shown in Figure \ref{fig:lpips_tradeoff}. We observe that a high prevalence of temporal redundancy indeed exists, indicated by the fact that after compressing $46\%$ of tokens, the decoded results still show high visual similarity (LPIPS $\le 0.05$, dashed line) compared with the original decoded video \citep{zhang2018perceptual}. This confirms that substantial temporal redundancy exists in the latent space, and we can take advantage of this property.

\section{Deriving Target Objective}
\label{appxpendix: Derive Target Objective}
Our goal is to show Eq. \ref{appendix:eqn:goal_app} is true in the Transformer architecture.
\begin{equation}
  \operatorname{MSA}(\mathcal{P}(x_t)) \approx \mathcal{P}(\operatorname{MSA}(x_t))  \implies     
    D(\mathcal{P}(x_t)) \approx \mathcal{P}(D(x_t))
\label{appendix:eqn:goal_app}
\end{equation}
where $x_t$ denotes the token sequence at time $t$, $\mathcal{P}$ represents the pruning operator, $D$ is the denoising network, and $\mathcal{R}$ denotes the recovery operator (reusing temporal predecessors).

Since the denoising network $D(\cdot)$ is a Diffusion Transformer composed of stacked attention blocks, ensuring equivalence at all block is a sufficient condition for global approximation (see Eqn. \ref{eqn:goal}):
\begin{equation}
    \operatorname{Block}(\mathcal{P}(x_t)) \approx \mathcal{P}(\operatorname{Block}(x_t))
    \label{eqn:block_equ_app}
\end{equation}
Within a standard Transformer block, the Feed-Forward Network (FFN) and Cross-Attention layers operate point-wise on the video tokens (\textit{i.e.}, $x_i' = f(x_i)$). Since the output of a token in these layers depends only on itself, they are unaffected by pruning.

However, the Multi-Head Self-Attention (MSA) layer introduces \textit{inter-token dependency}, where the calculation for a token $x_i$ depends on the entire sequence:
\begin{equation}
    x_i' = \operatorname{MSA}(x_0, x_1, \dots, x_N)_i
\end{equation}
Consequently, satisfying Eqn. \ref{eqn:block_equ_app} reduces to ensuring that the output of the self-attention layer remains the same under pruning:
\begin{equation}
    \operatorname{MSA}(\mathcal{P}(x_t)) \approx \mathcal{P}(\operatorname{MSA}(x_t))
    \label{eqn:msa_equivalence_app}
\end{equation}
\section{General Case for the Impact of I.I.D Noise}
\label{sec:noise_impact}
The quadratic noise term $\epsilon_i^T W \epsilon_j$, where $W = W_Q^T W_K$, $W_Q$ is the weight matrix for query, $W_K$ is the weight matrix for keys, and $\epsilon_i$ and $\epsilon_j$ are the Gaussian Noise added to the tokens, changes distribution from Gaussian distribution $\mathcal{N}$ to Chi squared $\chi^2$ distribution:
\begin{equation}
\epsilon_i^T W \epsilon_j \;\sim\;
\begin{cases}
\mathcal{N}(0, \|W\|_F^2) & \text{if } \epsilon_i \neq \epsilon_j \quad \text{(independent)} \\
\sum_{m=1}^D \lambda_m \chi^2_1 & \text{if } \epsilon_i = \epsilon_j \quad \text{(duplicated)}
\end{cases}
\label{appendix:eqn:noise_interact}
\end{equation}
 where $\lambda_m$ are the eigenvalues of the symmetric part of $W$, defined as $W_{\text{sym}} = \frac{1}{2}(W + W^T)$. Note that by the Central Limit Theorem, $\mathcal{N}(0, \|W\|_F^2)$ is an approximation for large token dimension $D$. The duplicated case introduces a positive bias ($\mathbb{E}[\epsilon_i^T W \epsilon_i] = \text{Tr}(W)$) and higher variance ($2\text{Tr}(W_{\text{sym}}^2)$). Since Transformer projection matrices are typically learned such that $W$ has a heavily positive trace (to ensure identical tokens attend to themselves), this bias is large and effectively inflates the attention weights on duplicated tokens.
\newpage
\section{Latent Inter-Frame Pruning and Restoration Full Algorithms}
\label{apx:algorithms}
\noindent
\begin{minipage}[t]{0.48\textwidth}
\begin{algorithm}[H]

\caption{Latent Inter-Frame Pruning}

\label{alg:LIF_pruning}

\begin{algorithmic}[1]

\STATE {\bfseries Input:} Video latent $  X = \{X_1, X_2, \dots, X_T\}  $, Temporal Stride $  k  $, Thresholds $  \tau_1, \tau_2  $

\STATE {\bfseries Output:} Keep Mask $  M_{\text{all}}  $
\STATE
\STATE \textbf{Function} \textsc{GetDiffMask}$  (A, B, \tau)  $:
\STATE \quad $  D \gets |A - B|  $
\STATE \quad $  M \gets \text{3D-GaussianAdaptiveThreshold}(D, \tau)  $
\STATE \quad \textbf{return} $  M  $
\STATE Initialize $  M_{\text{all}} \gets \emptyset  $
\FOR{$  t=0  $ {\bfseries to} $  T-1  $}
\STATE \textit{// Compute Short and Long-term Difference}
\IF{$  t = 0 $}
\STATE $  M_{\text{short}} \gets \text{all-true mask}  $  
\ELSE
\STATE $  M_{\text{short}} \gets \textsc{GetDiffMask}(X_t, X_{t-1}, \tau_1)  $
\ENDIF
\IF{$  t \leq k  $}
\STATE $  M_{\text{long}} \gets \text{all-true mask}  $
\ELSE
\STATE $  M_{\text{long}} \gets \textsc{GetDiffMask}(X_t, X_{t-k}, \tau_2)  $

\ENDIF \

\STATE \textit{// Combine and Smooth}

\STATE $  M_t \gets M_{\text{short}} \lor M_{\text{long}}  $

\STATE $  M_t \gets \text{3D-MedianBlur}(M_t)  $

\STATE $  M_t \gets \text{2D-Morphology}(M_t, Smoothing)  $

\STATE $  M_t \gets \text{3D-Dilation}(M_t)  $

\STATE Append $  M_t  $ to $  M_{\text{all}}  $

\ENDFOR

\STATE \textbf{return} $  M_{\text{all}}  $

\end{algorithmic}

\end{algorithm}
\end{minipage}
\hfill
\begin{minipage}[t]{0.48\textwidth}
\begin{algorithm}[H]

  \caption{Latent Patch Restoration}

  \label{alg:unprune}

  \begin{algorithmic}[1]

    \STATE {\bfseries Input:} Latent Patch $X$, Keep Mask $M$

    \STATE {\bfseries Output:} Restored Latent Patch $U$

    \STATE \textit{// Initialization}

    \STATE $U \gets \emptyset_{X.\text{shape}}$ \COMMENT{Initialize empty tensor}

    \STATE $U[M] \gets X$

    \STATE \textit{// Temporal Reconstruction Loop}

    \FOR{$t = 0$ {\bfseries to} $T-1$}
        \IF{$t = 0$}
            \STATE Continue  \COMMENT{First frame is always all true}
        \ELSE
            \STATE $Pruned \gets \neg M_t$ 
            \STATE $U_t[Pruned] \gets U_{t-1}[Pruned]$
        \ENDIF
    \ENDFOR

    \STATE \textbf{return} $U$

  \end{algorithmic}

\end{algorithm}
\end{minipage}

\paragraph{Latent Inter-Frame Pruning.}
Diffusion latent space contains temporal redundancy, which allows us to consider Inter-Latent Compression \citep{MPEG1991, choudhury2024rlt} to bypass calculating repeated tokens. The core idea of Latent Inter-frame Pruning (LIF) is to identify similar patches by comparing temporally consecutive patches with the same spatial location, as shown in Alg. \ref{alg:LIF_pruning}, Line 5:

\begin{equation}
\|X_{t} - X_{t+1}\|_1 < \tau 
\label{apx:eq:pruning_criterion},  
\end{equation}

Due to the high compression rate in the latent space, subtle movements in latent patches can yield small differences in Eqn.~\ref{apx:eq:pruning_criterion}, causing mispruning. In the restoration stage, we reuse previous tokens. This may result in glitches that propagate when decoded, degrading video quality.

To mitigate this, we integrate motion detection by calculating the difference in frames in Eqn.~\ref{apx:eq:pruning_criterion} to avoid mispruning tokens with subtle movement. The idea is to recognize that videos typically involve object-level movements rather than isolated pixel changes. As a result, we leverage temporal and spatial information from neighboring tokens to identify object movement. Specifically, in Alg.~\ref{alg:LIF_pruning}, L6 and L22-24, we apply a 3D adaptive Gaussian threshold to account for neighboring differences and median blurring when computing frame changes, followed by a closing morphological operator to eliminate isolated pruned tokens. We further dilate the mask to provide a margin around boundary tokens exhibiting minimal changes.

Additionally, we enhance the binary (keep) mask, which is True for kept tokens and False for pruned tokens, by incorporating both short-term (consecutive) and long-term temporal differences, as shown in Line 21. This dual-term design is critical for supporting attention recovery and preventing the violation of the I.I.D. noise assumption.
 
\section{Experimental Settings}
\label{appendix:experimental_settings}
We implement our pruning method on top of the Self-Forcing model~\citep{huang2025selfforcing}. Consistent with CausVid~\citep{yin2025causvid} and StreamV2V~\citep{liang2024looking}, we employ SDEdit~\citep{meng2022sdedit} for video-to-video translation. By default, we use a 4-step denoising schedule with an initial noise level of $t=400$ (out of $1000$). We use a Tiny autoencoder for encoding and decoding \citep{BoerBohan2025TAEHV}. The KV cache is trimmed for denoising and only preserves the most recent 6 frames due to the m-degree approximation. 

The pruning thresholds $\tau_1$ and $\tau_2$ (from Section~\ref{sec:token_pruning} and Eq.~\ref{eqn:second_mask}) are set to $0.15$ and $0.3$, respectively, resulting in an average of 32\% tokens pruned. All experiments were conducted on an NVIDIA A6000 GPU with a fixed random seed of 0 for reproducibility.

Following the evaluation methods of TokenFlow~\citep{tokenflow2023} and StreamV2V~\citep{liang2024looking}, we assess performance on object-centric videos from the DAVIS 2017 train-val dataset~\citep{davis2017dataset}. This dataset covers diverse subjects (\textit{e.g.}, humans, animals, cars, \textit{etc.}). The 102 video-prompt pairs used ranging from stylization to object swaps. Additionally, we also conduct evaluations on 120 video-prompt pairs on OpenHumanVid \citep{li2024openhumanvid}. We conduct a thorough comparison with state-of-the-art real-time (or low latency) V2V methods, including Self-Forcing~\citep{huang2025selfforcing}, StreamV2V~\citep{liang2024looking}, StreamDiffusion~\citep{kodaira2023streamdiffusion}, and ControlVideo~\citep{zhang2023controlvideo}, using their official implementations under default settings. To evaluate the performance, we rely on sixteen human participants to evaluate generated videos. Furthermore, we benchmark our approach against training-free pruning methods such as ToMe for SD~\citep{bolya2023tomesd}, Importance-based Token Merging~\citep{wu2025importancetome}, and IDM~\citep{fang2025attend}. In addition to qualitative observations, we perform quantitative evaluation by reporting Warp Error \citep{Lai-ECCV-2018} and VBench scores~\citep{huang2023vbench} for video quality, and throughput to measure latency.

\section{Webpage for Human Evaluation Test}
\label{appendix:human_eval_test}
\begin{figure}[h]
    \centering
    \includegraphics[width=1\linewidth]{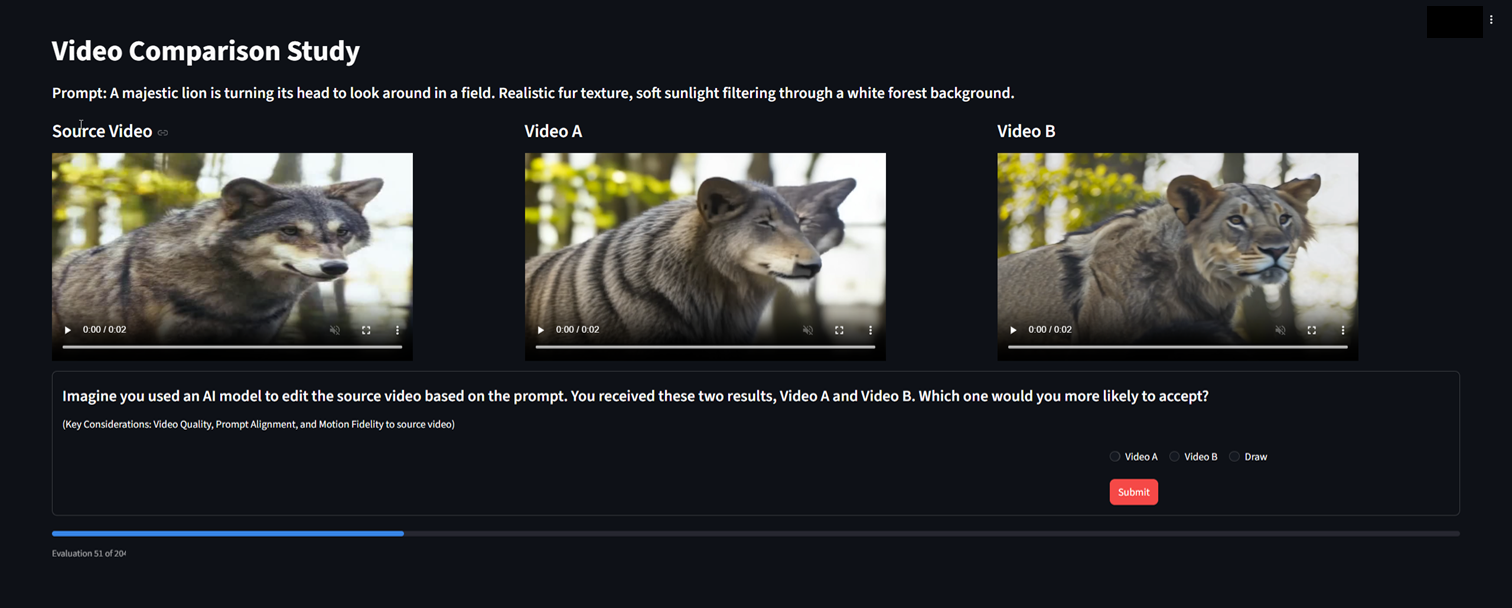}
    \caption{Webpage for performing human evaluation test.}
    \label{fig:human_eval_webpage_app}
\end{figure}

To validate the perceptual quality of our method, we conducted a user study comparing LIPAR against four baselines. Following TokenFlow~\citep{tokenflow2023} and StreamV2V~\citep{liang2024looking}, we use the DAVIS dataset~\citep{davis2017dataset}. We adopted a Two-Alternative Forced Choice (2AFC) protocol, where participants were presented with two videos side-by-side—one generated by our method and one by a baseline—and asked to select the better result, considering overall video quality (temporal consistency and frame quality), text-prompt alignment, and structural fidelity to the source video. The study involved 16 participants; each participant evaluated 25 randomly selected prompt pairs against all four baselines, resulting in 100 pairwise comparisons per participant. Figure \ref{fig:human_eval_webpage_app} displays the webpage used for conducting the human evaluation test.

\section{Further Discussion on Qualitative Comparison with Other Pruning Methods}
\label{appendix:compare pruning methods}
\begin{enumerate}
    \item \textbf{Throughput Difference:} Despite using identical pruning rates, LIPAR achieves significantly higher throughput (FPS) than the baselines. This is primarily because token merging methods incur substantial overhead by executing merge operations at regular intervals for excessive tokens. In contrast, LIPAR computes the pruning mask once per incoming frame, with an overhead of $0.65$--$0.69$\,ms per frame. Furthermore, while baseline token merging is restricted to the Self-Attention module (following~\citep{bolya2023tomesd}), our method applies pruning in an end-to-end manner across all layer components, maximizing acceleration.
    \item \textbf{Model Susceptibility to Token Merging:} The Self-Forcing model's causal attention mechanism is sensitive to token manipulation if positional encoding and noise correlations are not explicitly handled. Existing pruning methods did not address these factors, resulting in quality degradation. In contrast, we formulate conditions for preserving the pruned token value and handled these factors in LIPAR (see Section~\ref{sec:noise_aware_duplication}) to preserve quality. 
\end{enumerate}

\section{Latency VS. Tokens Number}
\label{appdx:latency_token_number}

\begin{figure}[h]
    \centering
    \includegraphics[width=0.75\linewidth]{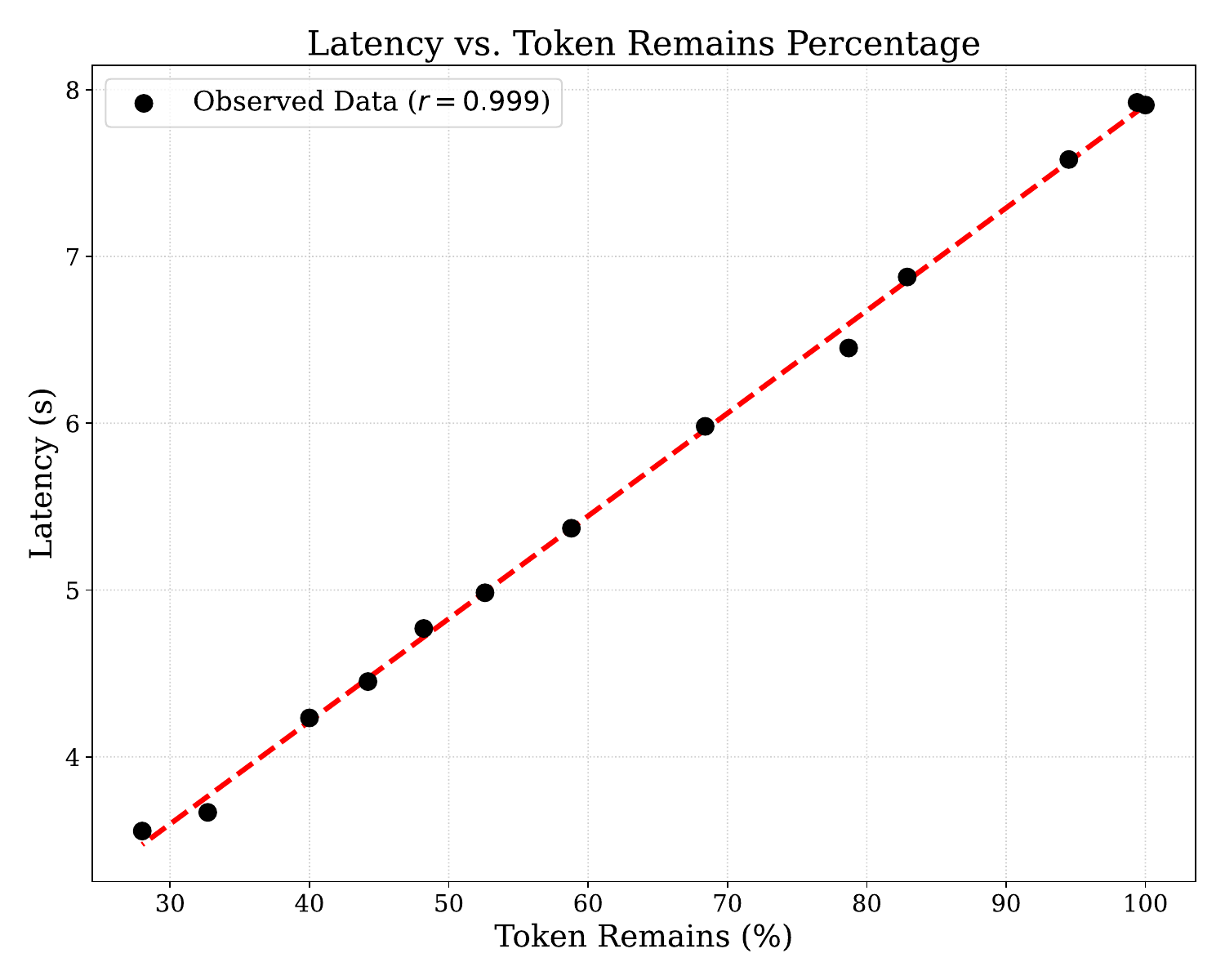}
    \caption{Latency on an NVIDIA A6000 GPU for generating a 4.5s video across varying token counts.}
    \label{fig:speed_token_remains}
\end{figure}

In Figure~\ref{fig:speed_token_remains}, we modulate the pruning rate by adjusting the threshold $\tau$ in LIPAR and measure the corresponding latency, with each data point averaged over ten runs. Consistent with Section~\ref{sec:attention_recovery}, we observe a strong linear correlation (Pearson $r=0.999$) between remaining tokens and latency, empirically verifying that LIPAR maintains $O(n)$ complexity even with Attention Recovery, where $n$ is the number of remaining tokens. This linearity also enables precise latency prediction before video editing, facilitating efficient GPU resource allocation.

\section{Detailed Long-Horizon Evaluation}
  \label{app:long_horizon_results}

  Table~\ref{tab:long_horizon_full} reports the full results for the long-horizon experiment in
  Section~\ref{sec:long_horizon}. We evaluate 12 one-minute video-prompt pairs from DH-FaceVid-1K~\citep{di2025facevid},
  each containing 960 frames at 16 FPS. Periodic refresh is enabled every six latent frames by disabling pruning in the
  final denoising step. We compute VBench scores~\citep{huang2023vbench} over twelve non-overlapping five-second windows
  and average the results across the video-prompt pairs. The dataset-averaged quality score in every window remains
  within $0.007$ of the twelve-window mean, and the final window's score ($0.857$) closely matches the first ($0.856$).
  A per-video linear fit gives a change of $+0.0002$ per ten seconds ($95\%$ CI: $[-0.0007,+0.0010]$), indicating no
  measurable quality decline over the evaluated duration.

  \begin{table}[ht]
      \centering
      \small
      \setlength{\tabcolsep}{5pt}
      \caption{Full long-horizon results with periodic refresh, averaged over 12 video-prompt pairs. Time denotes the
      start of each five-second window; the mean includes all twelve windows. Higher is better for all metrics.}
      \label{tab:long_horizon_full}
      \begin{tabular}{lccccccc}
          \toprule
          \textbf{Time (s)} & \textbf{Quality} & \textbf{Subj.} & \textbf{Backg.} & \textbf{Flicker} & \textbf{Motion} &
          \textbf{Aesth.} & \textbf{Img.} \\
          \midrule
          0    & 0.856 & 0.959 & 0.954 & 0.990 & 0.994 & 0.579 & 0.709 \\
          5    & 0.849 & 0.953 & 0.950 & 0.987 & 0.993 & 0.562 & 0.709 \\
          10   & 0.853 & 0.957 & 0.951 & 0.989 & 0.994 & 0.564 & 0.714 \\
          15   & 0.860 & 0.970 & 0.962 & 0.991 & 0.994 & 0.572 & 0.714 \\
          20   & 0.856 & 0.963 & 0.957 & 0.990 & 0.994 & 0.569 & 0.710 \\
          25   & 0.858 & 0.969 & 0.959 & 0.990 & 0.994 & 0.569 & 0.712 \\
          30   & 0.856 & 0.964 & 0.957 & 0.989 & 0.994 & 0.575 & 0.705 \\
          35   & 0.856 & 0.964 & 0.958 & 0.989 & 0.993 & 0.572 & 0.707 \\
          40   & 0.854 & 0.959 & 0.955 & 0.989 & 0.993 & 0.570 & 0.707 \\
          45   & 0.853 & 0.961 & 0.956 & 0.987 & 0.993 & 0.571 & 0.701 \\
          50   & 0.854 & 0.957 & 0.957 & 0.990 & 0.994 & 0.569 & 0.703 \\
          55   & 0.857 & 0.970 & 0.961 & 0.990 & 0.993 & 0.569 & 0.704 \\
          \midrule
          Mean & 0.855 & 0.962 & 0.956 & 0.989 & 0.994 & 0.570 & 0.708 \\
          \bottomrule
      \end{tabular}
  \end{table}

\newpage
\section*{NeurIPS Paper Checklist}

\begin{enumerate}

\item {\bf Claims}
    \item[] Question: Do the main claims made in the abstract and introduction accurately reflect the paper's contributions and scope?
    \item[] Answer: \answerYes{} 
    \item[] Justification: Yes, we summarized our contribution and scope in both abstract and introduction.

\item {\bf Limitations}
    \item[] Question: Does the paper discuss the limitations of the work performed by the authors?
    \item[] Answer: \answerYes{} 
    \item[] Justification: We mentioned the limitation of the scope in the conclusion and future work section of the paper.

\item {\bf Theory assumptions and proofs}
    \item[] Question: For each theoretical result, does the paper provide the full set of assumptions and a complete (and correct) proof?
    \item[] Answer:  \answerYes{} 
    \item[] Justification: The assumptions and derivations are given in Section~\ref{sec:proble_formulation} and Section~\ref{sec:attention_recovery}, with full derivations in Appendices C and D.

    \item {\bf Experimental result reproducibility}
    \item[] Question: Does the paper fully disclose all the information needed to reproduce the main experimental results of the paper to the extent that it affects the main claims and/or conclusions of the paper (regardless of whether the code and data are provided or not)?
    \item[] Answer: \answerYes{} 
    \item[] Justification: We include information for reproducibility in the experimental settings as well as supplementary materials.

\item {\bf Open access to data and code}
    \item[] Question: Does the paper provide open access to the data and code, with sufficient instructions to faithfully reproduce the main experimental results, as described in supplemental material?
    \item[] Answer:  \answerYes{} 
    \item[] Justification: The code is publicly available at \url{https://github.com/DennisMenn/lipar}.

\item {\bf Experimental setting/details}
    \item[] Question: Does the paper specify all the training and test details (e.g., data splits, hyperparameters, how they were chosen, type of optimizer) necessary to understand the results?
    \item[] Answer:  \answerYes{} 
    \item[] Justification: We did, please refer to the experimental settings.

\item {\bf Experiment statistical significance}
    \item[] Question: Does the paper report error bars suitably and correctly defined or other appropriate information about the statistical significance of the experiments?
    \item[] Answer: \answerNo{} 
    \item[] Justification: We use VBench and CLIP score following standard practice in the community.

\item {\bf Experiments compute resources}
    \item[] Question: For each experiment, does the paper provide sufficient information on the computer resources (type of compute workers, memory, time of execution) needed to reproduce the experiments?
    \item[] Answer: \answerYes{} 
    \item[] Justification: All experiments run on a single NVIDIA A6000 GPU (Section 6.1, Appendix F); per-frame time and peak GPU memory are reported in Table 6.
    
\item {\bf Code of ethics}
    \item[] Question: Does the research conducted in the paper conform, in every respect, with the NeurIPS Code of Ethics \url{https://neurips.cc/public/EthicsGuidelines}?
    \item[] Answer: \answerYes{} 
    \item[] Justification: We did follow Neurips Code of Ethics.

\item {\bf Broader impacts}
    \item[] Question: Does the paper discuss both potential positive societal impacts and negative societal impacts of the work performed?
    \item[] Answer: \answerYes{} 
    \item[] Justification: We have a section in the Appendix for broader impacts. 
    
\item {\bf Safeguards}
    \item[] Question: Does the paper describe safeguards that have been put in place for responsible release of data or models that have a high risk for misuse (e.g., pre-trained language models, image generators, or scraped datasets)?
    \item[] Answer: \answerNA{} 
    \item[] Justification: We do not identify such risks in our method.

\item {\bf Licenses for existing assets}
    \item[] Question: Are the creators or original owners of assets (e.g., code, data, models), used in the paper, properly credited and are the license and terms of use explicitly mentioned and properly respected?
    \item[] Answer:  \answerYes{} 
    \item[] Justification: We properly credit the owner of dataset, code, and models used in our research.

\item {\bf New assets}
    \item[] Question: Are new assets introduced in the paper well documented and is the documentation provided alongside the assets?
    \item[] Answer: \answerYes{} 
    \item[] Justification: The code is released at \url{https://github.com/DennisMenn/lipar}.

\item {\bf Crowdsourcing and research with human subjects}
    \item[] Question: For crowdsourcing experiments and research with human subjects, does the paper include the full text of instructions given to participants and screenshots, if applicable, as well as details about compensation (if any)? 
    \item[] Answer: \answerYes{} 
    \item[] Justification: Appendix G describes the instructions given to participants and Figure 11 shows the evaluation webpage. Participants were volunteers.

\item {\bf Institutional review board (IRB) approvals or equivalent for research with human subjects}
    \item[] Question: Does the paper describe potential risks incurred by study participants, whether such risks were disclosed to the subjects, and whether Institutional Review Board (IRB) approvals (or an equivalent approval/review based on the requirements of your country or institution) were obtained?
    \item[] Answer: \answerNo{} 
    \item[] Justification: No IRB approval was obtained. The study only asked volunteers to compare pairs of generated videos, which poses minimal risk to participants.

\item {\bf Declaration of LLM usage}
    \item[] Question: Does the paper describe the usage of LLMs if it is an important, original, or non-standard component of the core methods in this research? Note that if the LLM is used only for writing, editing, or formatting purposes and does \emph{not} impact the core methodology, scientific rigor, or originality of the research, declaration is not required.
    \item[] Answer: \answerNA{} 
    \item[] Justification: LLM does not play a critical role.

\end{enumerate}

\end{document}